\documentclass[conference]{IEEEtran}
\IEEEoverridecommandlockouts
\usepackage[nocompress]{cite}
\usepackage{amsmath,amssymb,amsfonts}
\usepackage{graphicx}
\usepackage{textcomp}
\usepackage{xcolor}
\usepackage[hidelinks]{hyperref}

\usepackage{amsmath, amsfonts, amssymb, amsthm}
\usepackage{verbatim}
\usepackage{url}
\usepackage{color}
\usepackage[noend]{algpseudocode}
\usepackage{multirow}

\usepackage{siunitx}
\ifx\BlackBox\undefined
\newcommand{\BlackBox}{\rule{1.5ex}{1.5ex}}  
\fi
\ifx\proof\undefined
\newenvironment{proof}{\par\noindent{\bf Proof\ }}{\hfill\BlackBox\\[2mm]}
\fi

\newcommand\shortsection[1]{\vspace{6pt}{\noindent\bf #1.}}

\theoremstyle{definition}

\makeatletter
\def\url@leostyle{%
  \@ifundefined{selectfont}{\def\UrlFont{\sf}}{\def\UrlFont{\small\sffamily}}}
\makeatother
\makeatletter
\def\url@beostyle{%
  \@ifundefined{selectfont}{\def\UrlFont{\sf}}{\def\UrlFont{\scriptsize\sffamily}}}
\makeatother

\newcommand{\redacted}[1]{{\emph{$<$Redacted for Anonymity$>$}}}

\title{Dissecting Distribution Inference}
\author{
}
\author{\IEEEauthorblockN{Anshuman Suri}
\IEEEauthorblockA{University of Virginia\\
\textit{anshuman@virginia.edu}}
\and
\IEEEauthorblockN{Yifu Lu}
\IEEEauthorblockA{University of Michigan \\
\textit{yifulu@umich.edu}}
\and
\IEEEauthorblockN{Yanjin Chen}
\IEEEauthorblockA{University of Virginia\\
\textit{yc2kg@virginia.edu}}
\and
\IEEEauthorblockN{David Evans}
\IEEEauthorblockA{University of Virginia\\
\textit{evans@virginia.edu}}
}

\usepackage{subcaption}
\usepackage{nicefrac}       
\usepackage{microtype}
\usepackage{booktabs}       
\usepackage{caption}
\theoremstyle{definition}
\usepackage{dashbox}
\usepackage[
    n,
    operators,
    advantage,
    sets,
    adversary,
    landau,
    probability,
    notions,    
    logic,
    ff,
    mm,
    primitives,
    events,
    complexity,
    asymptotics,
    keys]{cryptocode}
\usepackage{caption}
\usepackage{cleveref}
\usepackage{multirow}
\usepackage{diagbox}

\usepackage{tablefootnote}

\def\BibTeX{{\rm B\kern-.05em{\sc i\kern-.025em b}\kern-.08em
    T\kern-.1667em\lower.7ex\hbox{E}\kern-.125emX}}

\begin{document}

\newcommand{\anote}[1]{[\textcolor{blue}{\textbf{Anshuman:} #1}]}
\newcommand{\yanote}[1]{[\textcolor{blue}{\textbf{Yanjin:} #1}]}
\newcommand{\yianote}[1]{[\textcolor{blue}{\textbf{Yifu:} #1}]}
\newcommand{\fixnum}[1]{\textcolor{red}{#1}}
\newcommand{\fixed}[1]{#1}
\newcommand{\changed}[1]{#1}
\newcommand{\boneage}{RSNA Bone Age}
\newcommand{\graph}{ogbn-arxiv}
\newcommand{\census}{Census}
\newcommand{\censusn}{Census19}
\newcommand{\texas}{Texas-100X}
\newcommand{\celeba}{CelebA}
\newcommand{\botnet}{Chord}
\newcommand{\ltest}{Loss Test}
\newcommand{\ttest}{Threshold Test}
\newcommand{\ttabbr}{TT}
\newcommand{\kldtest}{KL Divergence Attack}
\newcommand{\kldabbr}{KL}
\newcommand{\pinmc}{Permutation Invariant Networks}
\newcommand{\pinabbr}{PIN}
\newcommand{\amc}{\changed{ERROR}}
\newcommand{\amcabbr}{\changed{ERROR}}
\newcommand{\zhangatt}{ZTO}
\newcommand{\combatt}{ERROR}
\newcommand{\gpptt}{Generative Per-Point Threshold Test}
\newcommand{\gzero}{$\mathcal{G}_0$}
\newcommand{\gone}{$\mathcal{G}_1$}
\newcommand{\neff}{\ensuremath{n_\textrm{leaked}}}

\newcommand{\gphsize}{0.95}

\makeatletter
\def\url@leostyle{%
  \@ifundefined{selectfont}{\def\UrlFont{\sf}}{\def\UrlFont{\small\sffamily}}}
\makeatother

\makeatletter
\def\url@sleostyle{%
  \@ifundefined{selectfont}{\def\UrlFont{\sf}}{\def\UrlFont{\footnotesize\sffamily}}}
\makeatother

\urlstyle{leo}

\maketitle

\begin{abstract}
A distribution inference attack aims to infer statistical properties of data used to train machine learning models. These attacks are sometimes surprisingly potent, but the factors that impact distribution inference risk are not well understood and demonstrated attacks often rely on strong and unrealistic assumptions such as full knowledge of training environments even in supposedly black-box threat scenarios.
To improve understanding of distribution inference risks, we develop a new black-box attack
that even outperforms the best known white-box attack in most settings. Using this new attack, we evaluate distribution inference risk while relaxing a variety of assumptions about the adversary's knowledge under black-box access, like known model architectures and label-only access. Finally, we evaluate the effectiveness of previously proposed defenses and introduce new defenses. 
We find that although noise-based defenses appear to be ineffective, a simple re-sampling defense can be highly effective.
\end{abstract}


\thispagestyle{plain}
\pagestyle{plain}

\section{Introduction} \label{sec:intro}

Machine learning models are susceptible to several disclosure risks, including leaking sensitive information related to training data.
Distribution inference considers what a model reveals about its entire underlying training data, in contrast with inference attacks that focus on individual records like membership inference and memorization attacks. Initial work focuses on inferring ratios of data with a particular attribute, referring to the attacks as “property inference”. Examples include inferring the accent of speakers in voice recognition models~\cite{ateniese2015hacking}, targeting ratios of characteristics like gender labels~\cite{ganju2018property}, and estimating sentiment across email datasets~\cite{mahloujifar2022property}.
More recently, there have been attempts to extend these attacks to properties beyond ratios, such as predicting graph density~\cite{zhang2021inference}, node/edge properties of groups within graphs~\cite{Wang2022GroupPI}, and direct regression over graph mean-degree~\cite{suri2022formalizing} with successful inference adversaries with just black-box access.

State-of-the-art distribution inference attacks achieve non-trivial distinguishing accuracies~\cite{ganju2018property, suri2022formalizing, zhang2021leakage} and thus pose a privacy risk, but the actual amount of leakage achieved is often minimal.
Leakage varies significantly across different datasets, but for most settings the best current attacks leak no more information than what one or two samples from the distribution would reveal~\cite{suri2022formalizing}. This is enough to have a significant advantage in distinguishing highly dissimilar distributions, but seems unlikely to pose a serious privacy risk in most cases. 

\shortsection{Contributions}
We advance the understanding of distribution inference risk on several fronts including an improved attack, analysis of risk, and development and evaluation of defenses.

We introduce a new black-box attack, the \kldtest, that uses distributional similarity in predictions (\autoref{sec:bb_attacks}) and substantially outperform the current state-of-the-art (\autoref{sec:experiments}), increasing previous estimates of inference leakage for various datasets in the literature. Surprisingly, we find that in most settings our black-box \kldtest\ is more effective than the best known white-box attack.
We also evaluate the black-box attacks in more realistic settings where the adversary does not have as much information as is typically assumed in inference experiments (\autoref{sec:training_envs}).

We evaluate the impact of different model architectures (\autoref{sec:diff_architectures}), the lack of common feature extractors (\autoref{sec:same_feature_extractor}), and relaxing the assumption of prediction probabilities to label-only access settings (\autoref{sec:preds_access}).
Our experiments find large variances in inference risk due to relaxing assumptions about model architectures and feature extractors, but demonstrate that attacks can be effective in label-only settings.

\autoref{sec:defenses} evaluates defenses, including both previously proposed ones and new ideas. Most privacy-related defenses for machine learning involve adding noise at some stage of the training; for instance, at the gradient-level in differentially private training~\cite{abadi2016deep}, or at the data level with most implementations of adversarial training~\cite{madry2018towards}. Our experiments find that these noise-based defenses provide little mitigation against distribution inference (\autoref{sec:noise_defenses}).
At some level, this is unsurprising since these defenses are designed to protect individual training records, not distributional properties, and we show a connection between how well a model generalizes to the task distribution and its susceptibility for distribution inference leakage (\autoref{sec:generalization}). We then develop a simple and inexpensive mitigation based on data re-sampling (Section~\ref{sec:rebalance}) which can protect against distribution inference in most cases where an adversary knows the statistical property to protect.

\section{Attacks} \label{sec:attacks}

Before introducing our attacks, we summarize the formal definition of distribution inference we use (\autoref{sec:formalize_di}). Then, we describe previous attacks and our new attack in the black-box (\autoref{sec:bb_attacks}) setting, and previous attacks in the white-box (\autoref{sec:amc}) setting.

\subsection{Defining Distribution Inference} \label{sec:formalize_di}

The goal of distribution inference is to infer sensitive properties of the distribution used to train a machine learning model given some level of access to that model. Mahloujifar et al.~\cite{mahloujifar2022property} formalized a notion of distribution inference that can capture the proportion of a dataset satisfying some property.
In this setting, each data-point can either have an attribute value (linked to the property, such as 'female') equal to zero or one, and sampling data from $\mathcal{D}_{+}$ and $\mathcal{D}_{-}$ corresponds to data with this attribute set to zero or one respectively. The two distributions to distinguish, then, correspond to a mixture of these distributions for some $\alpha$ corresponding to the probability of sampling a record from $\mathcal{D}_{+}$:
$$
    \alpha\cdot\mathcal{D}_{+} + (1-\alpha)\cdot\mathcal{D}_{-}
$$
Using this definition, the authors focus on the task of distinguishing between distributions with different $\alpha$ values, like ones with 40\% ($\alpha=0.4$) and 50\% ($\alpha=0.5$) females.
Suri and Evans~\cite{suri2022formalizing} generalize this notion by using distribution transformation functions \gzero\ and \gone\ to transform an underlying distribution $\mathcal{D}$ (which essentially corresponds to domain knowledge, such as the distribution of face images), instead of explicitly assuming binary attributes and the two distributions being mixtures with different $\alpha$ values. When considering proportional properties (related to $\alpha$, as described above), the transformers can correspond to sampling records with a given attribute value for any two valid probabilities, $\alpha_0$ and $\alpha_1$. 
We follow the formalization in Suri and Evans~\cite{suri2022formalizing}, since it is more generic and better captures assumptions like access to data from some underlying distribution $\mathcal{D}$. The adversary's task in this setting thus corresponds to distinguishing models trained on data from \gzero($\mathcal{D}$) and \gone($\mathcal{D}$), given access to a dataset sampled from $\mathcal{D}$ and some level of access to the trained model.

\subsection{Black-Box Attacks} \label{sec:bb_attacks}

Black-box attacks assume the adversary has the ability to submit inputs to the trained model and observe the response but does not have direct access to the model. 
In addition, the adversary has access to some representative data from some distribution $\mathcal{D}$, and seeks to infer which of the transformations, \gzero\ or \gone, corresponds to the victim's training distribution. 
Using knowledge of $\mathcal{D}$ and the transformation functions \gzero\ and \gone, the adversary is able to train shadow models locally.
Knowledge of the candidate distributions is necessary to be able to distinguish between them, and it is reasonable to assume an adversary with enough  computational resources to train models locally. Most research assumes that the victim and adversary use the same model architecture (e.g., \cite{zhang2021leakage, melis2019exploiting}), and that the adversary has access to
model prediction confidence vectors (e.g., \cite{zhang2021leakage, xu2019detecting, suri2022subject}). In \autoref{sec:adversaryknowledge}, we consider settings where the adversary has less knowledge of the victim model and the model API only outputs labels. 
Next, we review previous black-box distribution inference attacks, and then introduce our new \kldtest. 

\shortsection{Prior Work}
Zhang et al.~\cite{zhang2021leakage} propose meta-classifier attacks that use probability vectors from models for a specific set of query points. Similar ideas are explored in related tasks~\cite{xu2019detecting}. The attack works by collecting model predictions for a fixed set of query points (chosen at random); using local shadow models to train a meta-classifier on these concatenated predictions, and finally generating predictions for unseen models using the meta-classifier.
Suri and Evans~\cite{suri2022formalizing} propose the \ltest\ and \ttest\ attacks that compare model accuracies on candidate distributions to predict training distributions. The \ttest\ performs best of these: it uses locally trained models to derive a threshold on observed accuracy on a given data sample, which is then used to predict the training distribution of a model. This attack yields non-trivial inference accuracies in many cases but falls short of the white-box attacks by huge margins for most settings. These attacks have also been extended to settings where active adversaries that can poison the victim's training data~\cite{Chaudhari2022SNAPEE}. The only previous attacks designed to work with label-only predictions are by Juarez et. al.~\cite{Jurez2022BlackBoxAF} that performs a statistical test based on attribute-wise model performance, and  Mahloujifar et al.'s attack in the setting of active adversaries~\cite{mahloujifar2022property}.

\shortsection{\kldtest\ (\kldabbr)}
Recent work by Hartley et al.~\cite{hartley2022measuring} demonstrates how the presence of unique features, even if present in one training record, can impact output probability distributions. Motivated by their use of KL divergence to differentiate between the two scenarios (instance present or not), we propose an attack that compares the KL divergence in output probabilities of the victim model using local models.

The adversary prepares by training a collection of local models $\{M^1_0, M^2_0, ... M^1_1, M^2_1, ...\}$, where $M^i_0$ and $M^i_1$ (for some $i$) denote models from training distributions \gzero($\mathcal{D}$), \gone($\mathcal{D}$) respectively.
Let $X$ denote some data randomly sampled by the adversary from the distributions \gzero($\mathcal{D}$) and \gone($\mathcal{D}$), with an equal number of samples ($|X|/2$) from both distributions. We first define a way to estimate the KL-Divergence between two models using predictions:
\begin{align}
    \mathbb{E}[D_{KL}(N\parallel M)] = \mathbb{E}_{x \in X}\Bigg[\sum_{c \in \mathcal{C}} N(x)_c\log\bigg(\frac{N(x)_c}{M(x)_c}\bigg)\Bigg] \label{eq:kl_main_eq}
\end{align}
where $M(x)_c$ corresponds to the prediction probability corresponding to class $c$ (out of all classes $\mathcal{C}$) for some point $x$ for model $M$, and the expectation $\mathbb{E}[]$ is taken over the adversary's data $X$. We use the same data $X$ in computing KL-Divergence values.
Next, the adversary defines a ``weighted vote" for a pair of models $(N, P)$ with respect to $M$:
\begin{align}
    \lambda(M, N, P) = \mathbb{E}[D_{KL}(N\parallel M)] - \mathbb{E}[D_{KL}(P\parallel M)]. \label{eq:weight_vote}
\end{align}
A positive quantity $\lambda(M, N, P)$ thus indicates that the model $M$ has its predictions distributed closer to $P$ than $N$, since a lower KL-divergence between distributions indicates higher similarity. Using its collection of local models trained on the two candidate distributions, the adversary then computes and aggregates this ``weighted vote" across all pairs of its local models $(M^i_0, M^j_1)$:
\begin{align}
    \hat{b} = \mathbb{I}\bigg[\sum_i\sum_j \lambda(M, M^i_0, M^j_1) > 0\bigg] \label{eq:kdtest}
\end{align}
The rule above thus effectively checks all its pairs of local models and compares similarities in prediction distributions with a given victim model. Since the core idea here is to compare distributions of model predictions, other metrics to compare distributions, like Jensen-Shannon Divergence, or TV Distance, can be used instead of KL-Divergence.

\subsection{White-Box Attacks} \label{sec:amc}

In the white-box setting, the adversary additionally has direct access to the victim's model including its trained parameters. Although this access model assumes a stronger adversary, it is a realistic adversary for many scenarios, like when models are deployed on client devices.
It is also useful in two ways: 1) gauging the extent of inference leakage, helping bound risk and understand it better, and 2) studying patterns and trends across properties and models to help better understand distribution risk and come closer to inventing effective defenses. 

\shortsection{Prior Work}
The main previous white-box distribution inference attacks are based on permutation-invariant networks~\cite{ganju2018property}. These attacks assume that information related to the training distribution can be somehow extracted from trained model parameters. They look at model parameters across all layers of a multi-layer perceptron to generate a feature representation for the entire model that is insensitive to arbitrary reorderings of neurons. The method works by constructing feature representations for each layer using learnable parameters (with prior layers as context) using shadow models, and trains the meta-classifier via back-propagation, using labels indicating which training distribution the shadow models correspond to.
These attacks, originally designed for networks with linear layers, have been extended to support convolutional layers~\cite{suri2022formalizing}.
Other distribution inference attacks in the literature follow a similar meta-classifier approach: using parameter extraction for support vector machines~\cite{ateniese2015hacking}, using model gradients~\cite{melis2019exploiting}, or intermediate node embeddings in graph neural networks~\cite{Wang2022GroupPI}.

\section{Results} \label{sec:experiments}
We evaluate our proposed attacks on several datasets, including both established benchmarks used in prior work and new configurations and property-task combinations previously unexplored in the literature. Code for reproducing our experiments is available at \url{https://github.com/iamgroot42/dissecting_distribution_inference}.

Our new attacks are significantly more potent than the previous state-of-the-art.
Our \kldtest\ (\kldabbr) outperforms all previous black-box attacks by huge margins (\autoref{sec:first_results}). Even more interestingly, the \kldtest, with only black-box access, outperforms \pinmc\ (\pinabbr)\ by a large margin in nearly all settings. We study trends between the correlation of the task and property, and its impact on inference risk (\autoref{sec:correlation}).

\begin{table*}
\centering
\scalebox{0.90}{
\begin{tabular}{c c ccc c@{\hskip 0.3in} ccc c}
 \toprule
 \multirow{3}{*}{\textbf{Dataset}} & \multirow{3}{*}{\textbf{Task/Property}} & \multicolumn{4}{c}{\bf Distinguishing accuracy (\neff) for \boldmath $\alpha_1=0.2$} & \multicolumn{4}{c}{\bf Mean distinguishing accuracy (\neff)} \\
 & & \multicolumn{3}{c}{\textbf{Black-Box}} & \textbf{White-Box} & \multicolumn{3}{c}{\textbf{Black-Box }} & \textbf{White-Box} \\
  & & \ttabbr~\cite{suri2022formalizing} & \zhangatt~\cite{zhang2021leakage} &  \kldabbr & \pinabbr~\cite{ganju2018property} & \ttabbr~\cite{suri2022formalizing} & \zhangatt~\cite{zhang2021leakage} &  \kldabbr & \pinabbr~\cite{ganju2018property} \\
 \midrule
 \multirow{2}{*}{\censusn} & Income/Females & 50.0 ($<$0.1) & 53.4 ($<$0.1) & {\bf 89.8 (2.1)} & 78.6 (0.8) & 61.3 (0.9) & 54.4 ($<$0.1) & {\bf 82.5 (4.2)} & 81.0 (3.5)\\
 & Income/Whites & 53.2 ($<$0.1) & 52.6 ($<$0.1) & {\bf 92.4 (2.7)} & 74.2 (0.6) & 59.4 (0.7) & 54.9 ($<$0.1) & {\bf 83.7 (3.3)} & 75.4 (1.1) \\
 \midrule
 \multirow{3}{*}{\texas} & Procedure/Females & 50.0 ($<$0.1) & 50.0 ($<$0.1) & {\bf 89.3 (2.0)} & 50.0 ($<$0.1) & 51.2 ($<$0.1) & 51.6 ($<$0.1) & {\bf 82.5 (3.8)} & 51.3 ($<$0.1)\\
 & Procedure/Whites & 50.9 ($<$0.1) & 50.0 ($<$0.1) & {\bf 86.8 (1.7)} & 50.0 ($<$0.1) & 52.4 ($<$0.1) & 50.1 ($<$0.1) & {\bf 81.6 (3.7)} & 50.5 ($<$0.1)\\
 & Procedure/Hispanic & 50.0 ($<$0.1) & 50.0 ($<$0.1) & {\bf 78.4 (0.8)} & 50.0 ($<$0.1) & 50.0 ($<$0.1) & 50.0 ($<$0.1) & {\bf 82.4 (3.8)} & 50.1 ($<$0.1)\\
 \midrule
 \multirow{6}{*}{\celeba} & Mouth Open/Wavy & 52.0 ($<$0.1) & 51.8 ($<$0.1) & 56.8 ($<$0.1) & {\bf 92.0 (2.6)} & 50.6 ($<$0.1) & 52.3 ($<$0.1) & 62.1 ($<$0.1) & {\bf 86.1 (2.4)}\\
 & Smile/Females & 54.4 ($<$0.1) & 57.6 ($<$0.1) & {\bf 89.6 (2.1)} & 57.6 ($<$0.1) & 55.4 (0.1) & 60.9 (0.2) & {\bf 85.3 (3.2)} & 68.4 (0.5) \\
 & Gender/Young & 50.3 ($<$0.1) & 52.6 ($<$0.1) & {\bf 86.4 (1.6)} & 81.0 (1.0) & 52.9 ($<$0.1) & 55.5 (0.1) & {\bf 86.3 (2.5)} & 81.2 (1.5)\\
 & Mouth Open/Cheekbones & {50.0 ($<$0.1)} & {50.0 ($<$0.1)} & {84.6 (1.4)} & {\bf 95.8 (3.9)} & {50.1 ($<$0.1)} & {56.2 (0.1)} & {76.7 (1.4)} & {\bf 88.6 (3.0)} \\
 \midrule
 RSNA & Age/Females & 90.0 (2.2) & 95.4 (3.7) & {\bf 99.9 (20.1)} & 99.4 (7.9) & 64.0 (0.5) & 77.9 (1.6) & {\bf 94.5 (12.1)} & 95.2 (10.2)\\
 Bone Age & Females/Age & 95.7 (3.8) & 99.4 (7.9) & {\bf 99.9 (20.1)} & 66.0 (0.2) & 68.5 (1.0) & 78.5 (3.3) & {\bf 99.8 (22.6)} & 75.2 (8.4)\\
 \midrule
 \multirow2{*}{\graph} & Node classification/ & \multirow2{*}{50.0 ($<$0.1)} & \multirow2{*}{50.0 ($<$0.1)} & \multirow2{*}{\bf 99.9 (58.5)} & \multirow2{*}{87.4 (5.1)} & \multirow2{*}{50.1 ($<$0.1)} & \multirow2{*}{55.4 (6.2)} & \multirow2{*}{\bf 92.6 (182.5)} & \multirow2{*}{71.9 (11.7)}\\
 & Mean Degree \\
 \bottomrule
\end{tabular}
}
\caption{Effectiveness of inference attacks. 
We show results for our KL Divergence Attack (KL) and three prior attacks: \ttest\ (\ttabbr)~\cite{suri2022formalizing}, \zhangatt~\cite{zhang2021leakage}, and \pinmc\ (\pinabbr)~\cite{ganju2018property}. 
For the classifiers, the first set of results shows the attack's ability to distinguish between models trained on training sets where the proportion of the property is either $\alpha_0 = 0.5$ or $\alpha_1 = 0.2$ as an accuracy percentage, with corresponding  \neff\ values in parentheses.
The second set of results shows the mean distinguishing accuracies (\%) (with corresponding \neff\ values) between $\alpha_0 = 0.5$ and a set of varying $\alpha_1$ values ($0.0, 0.1, 0.2, 0.3, 0.4, 0.6, 0.7, 0.8, 0.9, 0.1$). For the graph datasets used for \graph, for the first set of results we use $\alpha_0 = 13$ are $\alpha_1 = 10$ as the two distributions; for the second set, we vary the mean node degree as the property, setting $\alpha_0=13$ and varying $\alpha_1$ in $[9, 10, 11, 12, 14, 15, 16, 17]$, and report the mean distinguishing accuracy (with mean \neff\ value).
For all of the results, for each $\alpha_1$ value, we compute the median over five trials. Mean accuracy (and  \neff) is then computed over the mean of these values for all $\alpha_1$ values.
For each setting, results for the most effective attack 
are bolded.
}
\label{table:datasets_desc}
\end{table*}

\subsection{Datasets and Models}

We evaluate our attacks on twelve task-property pairs across five datasets, summarized in \autoref{table:datasets_desc}. These were selected to directly compare results with previous works (\boneage, \graph, \celeba), to  study the impact of task-property correlation on inference risk (various property-task pairs for \celeba), and to include datasets representing real-world use-cases, like \censusn\ and \texas.

\textbf{\censusn}~\cite{census19git} is an updated and expanded version of the Adult \census\ dataset~\cite{bay2000uci} based on data from the US Census Bureau. It contains a mixture of numerical and categorical features, and the same prediction task. We focus on the ratio of whites (race) and females (sex) as properties, and use a two-layer feed-forward neural-network as the architecture.

\textbf{\texas}~\cite{texasgit} contains demographic and medical information for patients across hospitals. The original dataset uses 100 possible classes for surgical procedure prediction. We slightly modify the task and focus only on data from the top 20 classes, reducing it to a 20-class classification task. We focus on the ratio of whites (race), females (sex), and Hispanics (ethnicity) as properties, and use a two-layer feed-forward neural-network.

\textbf{\celeba}~\cite{liu2018large} contains collections of face images of celebrities. Each image is annotated with attributes. We use three different tasks: smile detection, gender prediction, and mouth-open prediction. We conduct experiments with a convolutional neural network trained from scratch for this dataset, with five convolutional layers and pooling layers followed by three linear layers, which is the smallest network we could find with reasonable task accuracy. For our experiments with feature extractors, we also conduct experiments where the adversary uses a pre-trained FaceNet~\cite{schroff2015facenet} model trained on the CASIA-WebFace~\cite{yi2014learning} dataset, with a two-layer network. It leads to a drop in performance (from $\sim92\%$ to $\sim82\%$), but the point of such an experiment is indeed to assess inference risk in more practical settings.

For the attack inference properties, we use the proportion of females (smile-detection task), old people (gender-prediction task), people with wavy hair (mouth-open-prediction task), and people with high cheekbones (mouth-open prediction task). These pairs are useful in comparing results with previous works, and also help cover a spectrum of different correlations between the task and property attributes.  

\textbf{\boneage}~\cite{halabi2019rsna} contains x-ray images of hands, and the standard task is to predict the patient's age in months. We convert the task to binary classification based on an age threshold ($>132$ months), and focus on the ratios of the females (available as metadata) as properties. We also consider a flipped scenario, where the task is to predict females, with the ratios of people below the age threshold as properties. We use a pre-trained DenseNet~\cite{huang2017densely} model for feature extraction, followed by a two-layer network for classification.
Similar to \celeba, we consider a setting where the adversary uses pre-trained feature extractor, while the victim trains models from scratch. Additionally, we also consider a setting where both the victim and adversary use the same feature extractor, but use different model architectures on top of the feature extractor

\textbf{\graph}~\cite{wang2020microsoft} is a directed graph of citations between computer science arXiv papers, with the task as predicted subject area categories (out of 40) using features extracted from paper documents. We infer the mean node-degree property of the graph, and use Graph Convolutional Networks~\cite{kipf2017semi}.

\subsection{Experimental Setup}

We build upon the experimental setup described in earlier works on distribution inference, using implementations and trained models provided by previous works~\cite{ganju2018property, suri2022formalizing}.
For each dataset, we create non-overlapping splits of data for the victim and adversary. For each dataset, we simulate $\mathcal{D}$ using the dataset itself. \gzero($\mathcal{D}$) is simulated by sampling from the dataset, such that the resulting distribution has $\alpha=0.5$ (or 13, for \graph), while \gone($\mathcal{D}$) is simulated for some $\alpha$ (which we vary across experiments).
We include the datasets and victim/adversary splits used in previous experiments, and include results on two new datasets.
For all of the experiments, we follow the processing pipeline described in Suri and Evans~\cite{suri2022formalizing} to obtain non-overlapping splits. Essentially, both parties sub-sample from their data splits (to achieve specific $\alpha$ values) with different random seeds, and train models on the sampled data.

We perform each experiment five times and report mean values with standard deviation in all of our experiments. Full experimental details are provided in the Appendix~\ref{app:exp_details}.

\shortsection{\kldtest}
Since using all pairs of adversary's models can be expensive, the attack uses a set fraction ($0.8$) of randomly chosen pairs to compute the expectation in \autoref{eq:kdtest}. We experiment with multiple values of this fraction, 
and observe comparable performance.
For each pair of local models, the attack collects the difference in KL values. These differences are then normalized across all differences observed for local models, after which the adversary uses voting-based aggregation to generate the final prediction. We also experimented with variants that do not include voting, as well as ones that flip the inputs to KL-Divergence computation (so $D_{KL}(B||A)$ instead of $D_{KL}(A||B)$), but find the current version to perform best.

\subsection{Results} \label{sec:first_results}\label{sec:bb_vs_wb}

Directly comparing distinguishing accuracies across different distributions can be misleading, since certain pairs of distributions might be easier to distinguish than others. To standardize comparisons, we use the \neff~\cite{suri2022formalizing} metric in our experiments. For the case of binary distinguishing, 
\neff\ can be computed using Theorem 4.2 in Suri and Evans~\cite{suri2022formalizing} as:

\begin{align}
\neff\ =
\frac{\log (4\omega(1-\omega))}{\log (\max\left(\frac{\min(\alpha_0, \alpha_1)}{\max(\alpha_0, \alpha_1)}, \frac{1 - \max(\alpha_0, \alpha_1)}{1 - \min(\alpha_0, \alpha_1)}\right))}. \label{eq:nleaked}
\end{align}
\noindent
The \neff\ value quantifies the amount of leakage observed by the attack, by connecting it with a Bayes-optimal classifier with access to that number of samples. It is equivalent to the adversary being able to draw samples from the training distribution and executing an optimally distinguishing statistical test.
The \neff\ values measure how much the inference attack leaks about the distributional property in a way that allows comparison across experiments, since the quantity helps capture the attacker's capabilities in a way that is independent of $\alpha_0$ and $\alpha_1$, thus enabling comparison across distributions. It can be interpreted as the minimum number of records an adversary would need to sample directly from the training distribution to expect to make equally accurate predictions. For instance, a distinguishing accuracy of 95\% between distributions with $\alpha_0, \alpha_1 = 0.5, 0.51$ is intuitively much harder than distinguishing between distributions with $\alpha_0, \alpha_1 = 0.5, 0.9$. This notion is exactly what \neff\ aims to capture, by standardizing comparison: \neff\ values in this case would correspond to $\approx 84$ and $\approx 3$ respectively.

Table~\ref{table:datasets_desc} summarizes the results of our distribution inference experiments. For each experiment, we report mean distinguishing accuracies between two distributions as well as the mean distinguishing accuracy across a set of different distributions, as detailed in the table caption. We also include corresponding \neff\ values.
For the classifiers, we vary $\alpha_1$ in $[0.0, 1.0]$ at intervals of $0.1$, and set $\alpha_0=0.5$ for the case of ratio-based properties, where a certain $\alpha$ value for a distribution means datasets sampled uniformly at random would have $\alpha$ fraction of the data with the property attribute 1, for e.g. ratio of females. The distinguishing accuracies thus correspond to predicting whether a model has the training distribution corresponding to $\alpha_0$ or $\alpha_1$ where random guessing would be 50\% accuracy, and perfect predictions would be 100\%.

The majority of experimental evaluations in the literature follow a binary classification scenario, where the adversary is supposed to distinguish between two potential training distributions with property values $\alpha_0$ and $\alpha_1$.
Although regression-based adversaries have been demonstrated~\cite{suri2022formalizing} as being strictly more powerful, they are much more computationally expensive and we leave evaluations in that scenario to future work.

\shortsection{Trends across datasets}
Inference leakage varies significantly across different datasets, with very little leakage for most cases in \texas, substantial leakage for \censusn, and exceptionally high leakage for the graph-based \graph\ dataset. The lack of virtually any inference risk in \texas\ is surprising, as the features contain the property label, and data splits are processed per hospital during generation, making the victim and adversary distributions highly similar. This difference in inference risk between \censusn\ and \texas, despite both being tabular datasets, reveals how just the nature of data (tabular, images) does not by itself determine inference risk and risk can vary unpredictably (at least based on current understanding) with aspects of the data. As previously observed by Suri et al.\ \cite{suri2022subject}, leakage is quite high for \boneage. Our new improved attacks identify vulnerable datasets, such as \censusn, that would have been considered low leakage risks using previous state-of-the-art attacks.

\shortsection{Comparing black-box attacks}
The \kldtest\ outperforms \ttest\ (\ttabbr) and the black-box attack by Zhang et. al.~\cite{zhang2021leakage} (which we refer to as \zhangatt) in all cases with large margins. Across all of the settings, \ttabbr\ and \zhangatt\ rarely achieve distinguishing accuracies above 75\% i.e. \neff\ above 1.0 (indicating that the observed leakage is less than what an adversary would learn by sampling a single record from the training distribution), whereas the \kldtest\ produces meaningful leakage for all of the datasets. The superiority of the \kldtest\ can be attributed to the use of pairs of local models and their trends (which grow in the order $\binom{n}{2}
$ for $n$ models), as opposed to using information from models in isolation in the other attacks.

\shortsection{Number of shadow models}
The black-box attacks use 50 shadow models per training distribution. We vary this number to 1) get an empirical lower bound on the number of shadow models required to achieve non-trivial leakage, and 2) study increase in information leakage with an increase in shadow models. Leakage is significant with only five shadow models per distribution in most cases, and improves with more local shadow models (\autoref{tab:num_models_adv}). 

\begin{table*}[tb]
\centering
\renewcommand{\arraystretch}{1.1}
\begin{tabular}{lcccccc} 
 \toprule
\multirow{2}{*}{\bf Dataset/Task} & \multicolumn{6}{c}{\bf Number of Shadow Models} \\
& 5 & 10 & 25 & 50 & 100 & 400\\
 \hline
 \censusn\ (Sex) & 73.0 (3.0) & 76.5 (3.3) & 81.3 (4.0) & 82.5 (4.2) & 86.4 (5.1) & 89.7 (6.6)\\
 \censusn\ (Race) & 77.2 (2.6) & 79.3 (2.9) & 81.3 (3.1) & 83.7 (3.3) & 84.2 (3.3) & 84.7 (3.4)\\
 \boneage\ (Age) & 97.3 (18.3) & 98.3 (19.0) & 99.3 (21.3) & 99.7 (22.6) & 99.7 (22.6) & 99.8 (22.8)\\
 \celeba\ (Sex) & 73.9 (1.1) & 78.6 (1.7) & 80.9 (2.4) & 85.3 (3.2) & 86.9 (3.9) & 89.2 (5.1)\\
 \bottomrule
\end{tabular}
\caption{Impact of varying the number of shadow models used by the adversary per distribution to launch its attacks. Values are mean distinguishing accuracies (\%) (with mean \neff\ in parentheses) for \kldtest (computed as described in Table~\ref{table:datasets_desc}). Even with only 5 models, the adversary is able to achieve considerable inference leakage.}
\label{tab:num_models_adv}
\end{table*}

\shortsection{White-box attacks}
The black-box \kldtest\ performs surprisingly well despite the weaker threat model, outperforming the best white-box attack in nearly all experimental settings.
Since an adversary in the white-box setting has access to more information than just the data and model predictions, it should be at least as powerful as a black-box adversary. 
We attribute the relative ineffectiveness of the white-box attacks to two main reasons. First, in \pinmc, model parameters are directly used as features for the meta-classifier, unlike comparisons in model prediction distributions in \kldtest. Secondly, white-box attacks have a larger feature space and learning meta-classifiers additionally requires learning to recognize relevant patterns in model parameters, a huge and complex data distribution. The black-box attacks, on the other hand, are agnostic to parameters in the victim model and thus much easier to scale, resulting in better performance. 

\subsection{Correlation} \label{sec:correlation}

The impact of correlation between the underlying task of a model and the property of its training distribution being inferred has been touched upon briefly
in the literature~\cite{zhang2021leakage}, but not studied extensively.
Intuition suggests there should be some positive relationship between inference risk with increasing task-property correlation, but prior studies do not evaluate inference risk across a range of property-task correlations. We carefully pick pairs of properties and tasks for the \celeba\ dataset, such that there is a good range of correlations. 

We conduct experiments with property correlations of $\approx 0$ (Mouth Slightly Open--Wavy Hair), $\approx 0.14$ (Smiling--Female), $\approx 0.28$ (Female--Young), and $\approx 0.42$ (Mouth Slightly Open--High Cheekbones).
Across this range of correlations, mean distinguishing accuracies (\neff\ values in parantheses) are 62.1\% ($<$0.1), 85.3\% (3.2), 86.3\% (2.5), 76.7\% (1.4) as correlation values increase. The lack of any clear trend between correlation and inference risk is consistent with observations in the literature around task-property correlation and inference risk~\cite{zhang2021leakage}. As observed, inference risk is non-trivial as long as the task-property correlation is non-zero. While the case of non-zero inference risk is obvious (with zero correlation, loss optimization would not use the property as an indicative feature), changes in inference risk with varying correlation values may be  tied to observed correlation versus actual causality, and methods for causal learning may help alleviate this inference risk~\cite{tople2020alleviating}.

\begin{figure}[ptb]
\centering
    \includegraphics[trim={0.6cm 0.7cm 0.5cm 0.5cm}, clip, width=\gphsize\columnwidth]{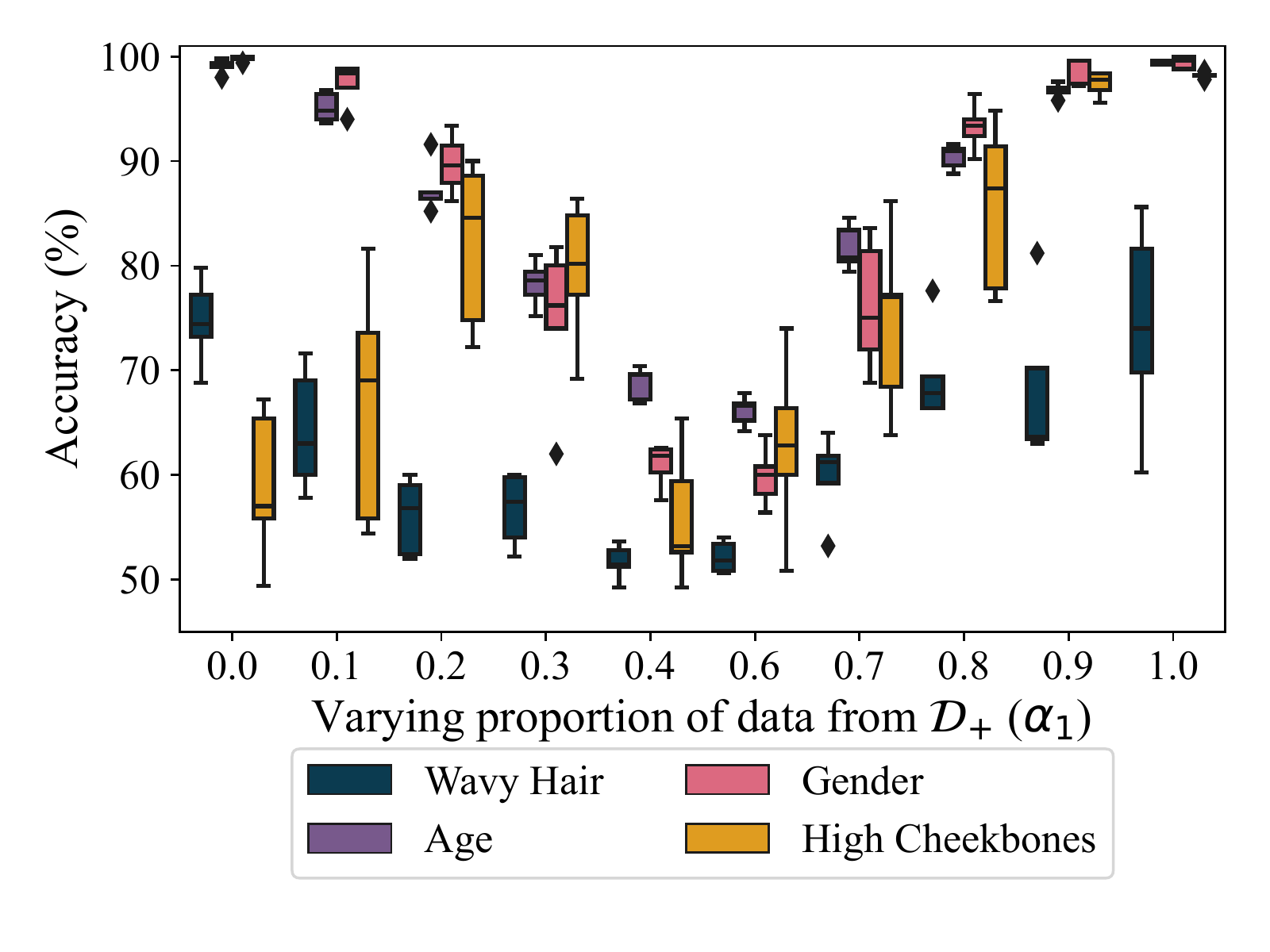}
\caption{Distinguishing accuracy for different task-property pairs for \celeba\ with varying correlation, for \kldtest.}
\label{fig:celeba_task_prop_pairs}
\end{figure}

We perform similar analyses for the \boneage\ dataset, where we flip the property and task. In this case, the correlation between the task and property remains the same, thus helping identify potential changes in inference leakage arising purely from the choice of the property itself.
While switching from Age--Females to Females--Age, we observe a huge bump in mean distinguishing accuracies (\neff\ values in parentheses): from 94.5\% ($12.1$) to 99.8\% ($22.6$).
Although the choice of property and task are expected to impact inference risk, our results suggest that this choice may be more relevant to evaluating inference risk than property-task correlation itself.

\section{Impact of Adversary's Knowledge} \label{sec:training_envs}\label{sec:adversaryknowledge}

Research on inference privacy typically considers threat models with one of two simplistic adversarial assumptions: white-box settings, where the adversary has full access to the model; and black-box settings, where the adversary has only API access to the model but receives full confidence vectors for each prediction and has complete knowledge of aspects of the training process and model architecture. The specific information available to an adversary in the black-box setting, however, may have a significant impact on inference risk. For instance, access to labeled data (for attacks) with prediction probabilities is often implicit, as is the use of the same model architectures and feature extractors between the victim and adversary. We study the impact of these common assumptions, and how relaxing them impacts inference risk. We measure impact on risk when the victim and adversary use different model architectures (Section~\ref{sec:diff_architectures}), do not share feature extractors (Section~\ref{sec:same_feature_extractor}), and when the available model API only provides label predictions  (Section~\ref{sec:preds_access}).
Inference risk is somewhat robust to differences in model architectures, as long as the victim and adversary's models have similar
capacity. The absence of shared feature extractors reduces inference risk significantly, but we find attacks can still succeed when only label predictions are available.

\subsection{Model Architecture} \label{sec:diff_architectures}

In the white-box setting an adversary can directly observe the target model's architecture, but in black-box settings it is unrealistic to assume the adversary knows the target model architecture. Likely model architectures may be limited in certain domains like image data, where the victim is likely to use a popular model architecture such as DenseNet~\cite{huang2017densely}. But a variety of models like random forests, support vector machines, and clustering-based classifiers can be used for tabular data and may even be picked by model trainers via automated tools~\cite{feurer2015efficient}.

Differences in victim and adversary architectures have not been previously explored, except by Mahloujifar et al.~\cite{mahloujifar2022property} for poisoning-based adversaries. In their setting, the victim and adversary can have different model architectures---the adversary uses logistic regression while the victim can use a variety of different feed-forward neural networks. They note a drop in inference risk with an increase in victim model complexity.
It is thus unclear if these trends are specific to the model architecture themselves. Additionally, the adversary's model architecture is kept the same, so they did not explore the potential for higher inference risk with better local models.

\begin{table*}[tb]
\centering
\begin{tabular}{lcccc} 
 \toprule
  \multirow{2}{*}{\bf Victim Model} & \multicolumn{4}{c}{\bf Adversary Model} \\
& RF & LR & MLP$_2$ & MLP$_3$\\
 \hline
 Random forest (RF) & 95.1 (12.0) & 78.9 (1.7) & 86.7 (5.4) & 85.6 (4.9) \\
 Linear regression (LR) & 93.2 (13.5) & 100.0 (25.9) & 76.4 (3.7) & 80.8 (5.4)\\
 Two-layer perceptron (MLP$_2$) & 69.7 (0.9) & 56.6 (0.3) & 82.5 (4.2) & 82.7 (4.3) \\
 Three-layer perceptron (MLP$_3$) & 69.3 (0.8) & 56.3 (0.3) & 82.2 (4.0) & 81.1 (3.8) \\
 \hline
\end{tabular}
\caption{Variation by model type. Each value is the observed mean distinguishing accuracy (\%) (with mean \neff\ in parentheses; measured as described in \autoref{table:datasets_desc}) of the \kldabbr\ attack for \censusn\ (Sex), for different combinations of model types for victim and adversary.}
\label{table:diff_models_censusn}
\end{table*}

To identify trends in inference risk with differences between architectures, we train multiple models with different architectures for both the victim and adversary. For \census\ (Gender), we try all possible combinations out of linear regression (LR), multi-layer perceptrons with two and three layers (MLP$_2$, MLP$_3$), and a random forest classifier (RF). We also consider using a two-layer perceptron (MLP$_2$) and a support vector machine (SVM) for the case of \boneage\ (Gender). For this experiment and the rest of this section, we report results with \kldtest.

We observe several interesting trends for \censusn\ while varying model types for the victim and adversary (Table~\ref{table:diff_models_censusn}). For instance, inference risk is significantly higher when the adversary uses models with learning capacity similar to the victim, like both using one of (MLP$_2$, MLP$_3$) or (RF, MLP). Concretely, mean distinguishing accuracy is 86.9\% (mean \neff=8.7) when learning capacities match, as opposed to 72.7\% (mean \neff=2.7) when learning capacities do not match.

Interestingly though, we also observe a sharp increase in inference risk when the victim uses models with low capacity, like linear regression and random forest instead of multi-layer perceptrons. For example, mean distinguishing accuracy is 72.6\% (mean \neff=2.3) when victim models have high learning capacity (MLP$_2$, MLP$_3$), but increases to 87.1\% (mean \neff=9.1) when the victim models have low learning capacity (RF, LR). These trends hint at possible connections between distribution inference risk and model learning capacity. 

\subsection{Feature Extractors} \label{sec:same_feature_extractor}
When dealing with high-dimensionality datasets and a scarcity of data, it is common to use techniques such as transfer learning~\cite{xie2018pre, wei2021finetuned} to boost model performance with reduced data and computational requirements. Using a pre-trained model for feature extraction should intuitively limit distribution-related privacy leakage, since there are fewer trainable parameters that can potentially contain revealing information. At the same time, fewer parameters also reduce the space of models, making it easier for an adversary to launch attacks. Even in a black-box setting, the adversary may be able to use the same feature extractor as the victim, either as a result of the adversary snooping and gaining information, or just by assuming the use of popular pre-trained models (like BERT~\cite{kenton2019bert}). This setting has been explored previously~\cite{ateniese2015hacking, ganju2018property, suri2022formalizing}, but the effect of the adversary using the same or different extraction models from the victim model has not been previously explored.

For \boneage\ (Sex), we consider two configurations: one where the victim and adversary use the same feature extractor, and another where the victim trains DenseNet~\cite{huang2017densely} models from scratch (CNN). For the first setting, we explore an SVM (FE+SVM) and a two-layer perceptron (FE+MLP$_2$).
There is a considerable drop in distinguishing accuracies (from 96.7\% to 91.2\%, i.e. \neff\ from 16 to  6) when the victim and adversary no longer share feature extractors (Table~\ref{table:diff_models_boneage}). For the settings where they do, we observe leakage to be highest for similar model architectures, and note a sharp increase when the victim uses an SVM. Nonetheless, inference risk stays sufficiently high. Interestingly, for the case where feature extractors are not shared, using a lower learning-complexity model (FE+SVM) seems to lead to higher leakage, than FE+MLP$_2$. While leakage is high in both cases, the increase can be explained by the chances of adversary's local models overfitting being less than that with an MLP.

For \celeba\ (Sex), we explore a setting where the adversary utilizes a feature extractor to train its models, while the victim trains CNNs from scratch. This setup represents a resource-constrained adversary who uses pre-trained models to lower computational and data requirements. We observe a similar diminishing of inference risk when a shared feature extractor is not available to the adversary, consistent with the RSNA Bone Age results. Compared to the scenario where the adversary uses the same model architecture as the victim without any pre-trained feature extractors, mean distinguishing accuracy drops from 85.3\% to 71.0\% (i.e., \neff\ from 3.2 to 0.5).

\begin{table}[tb]
\centering
\renewcommand{\arraystretch}{1.1}
\begin{tabular}{lcc} 
 \toprule
   \multirow{2}{*}{\bf Victim Model} & \multicolumn{2}{c}{\bf Adversary Model} \\
   & FE+MLP$_2$ & FE+SVM\\
 \midrule
 Feature extractor, perceptron (FE+MLP$_2$) & 94.5 (12.1) & 93.0 (9.0)\\
 Feature extractor, SVM (FE+SVM) & 99.5 (21.2) & 99.6 (21.7)\\
 DenseNet (CNN) & 88.0 (3.4) & 94.4 (8.5)\\
 \bottomrule
\end{tabular}
\caption{Mean distinguishing accuracies (\%) (with mean \neff\ values in parentheses) for \boneage\ (Sex), for different combinations of model types for victim and adversary (as computed in Table~\ref{table:datasets_desc}).}
\label{table:diff_models_boneage}
\end{table}
\begin{figure}[ptb]
\centering
    \includegraphics[trim={0.6cm 0.5cm 0.5cm 0.55cm}, clip,width=\gphsize\columnwidth]{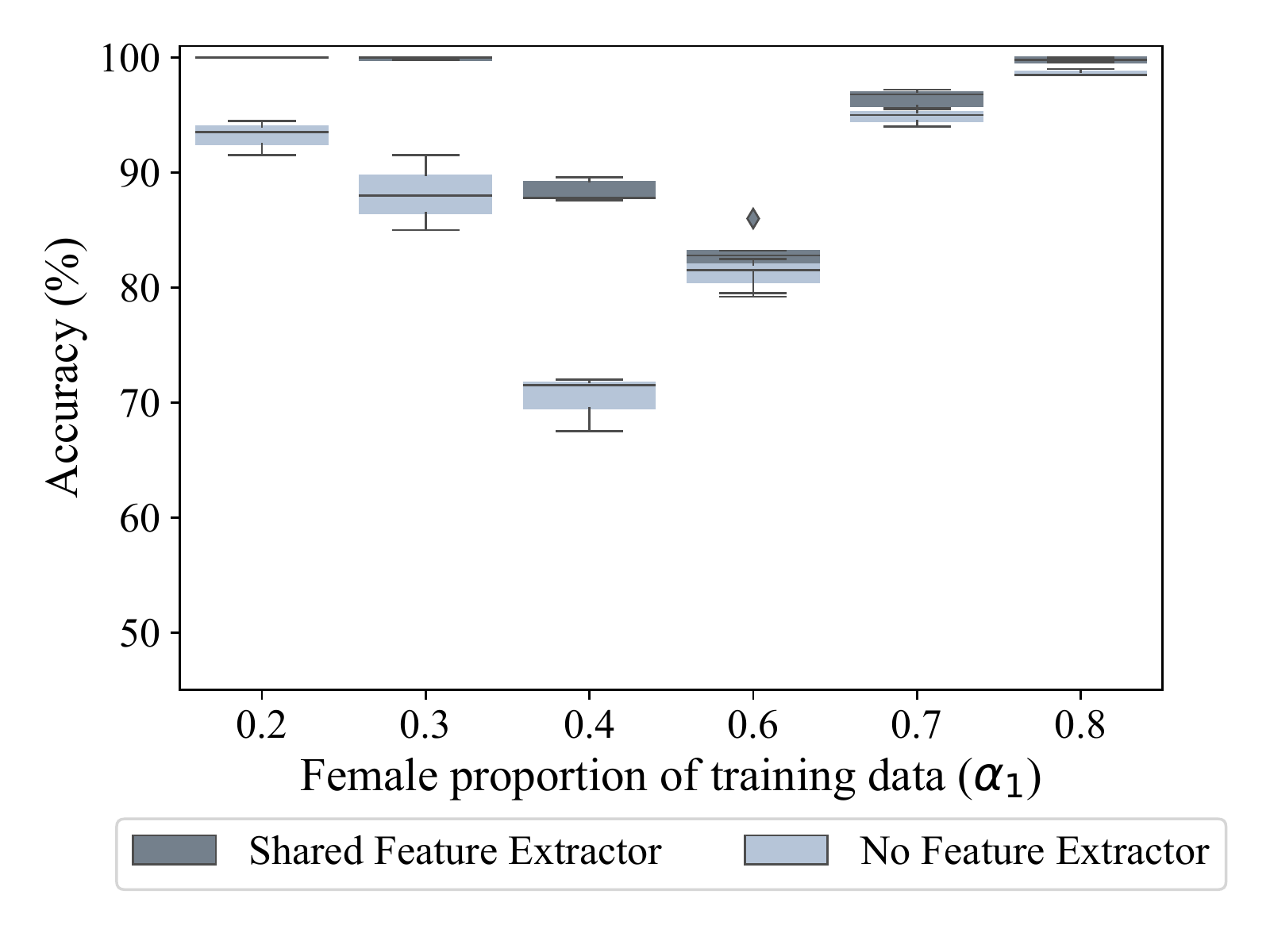}
\caption{Distinguishing accuracy for for the \kldtest\ for \boneage\ (Sex), when the adversary uses the same feature extractor as the victim, and when the victim does not use or share any pre-trained feature extractor. While there is an obvious drop in performance, inference risk still stays high.}
\label{fig:boneage_bb_full}
\end{figure}

\subsection{Label-Only Access}
\label{sec:preds_access}

Most black-box attacks in the literature related to distribution inference assume access to prediction confidence vectors. This is not an unreasonable assumption---many APIs return prediction scores, especially for top-k classes (for example \hyperlink{https://cloud.google.com/vision}{Google Vision API} and \hyperlink{https://www.clarifai.com/products/armada-ml-prediction}{ClarifAI Prediction API} return scaled confidence scores for the top 10 or 20 classes, respectively). It is unclear, however, what kind of performance drops to expect for distribution inference attacks in settings where the model's API only returns a label. The only previous works to explore distribution inference in the label-only setting are in the context of group distribution shift auditing~\cite{Jurez2022BlackBoxAF}, and active adversaries with poisoning capabilities~\cite{mahloujifar2022property}. 

With some straightforward modifications, \kldtest\ can be launched with access to just label predictions, with negligible drops in inference leakage in most cases.
The attack requires prediction confidence scores to compute the KL-divergence values. However, these scores are absent in the label-only setting, and the labels effectively correspond to confidence values of $0$ and $1$. This makes the KL computations (\autoref{eq:kl_main_eq}) invalid, since the log of $0$ or $1/0$ is undefined. To tackle this, we replace $0/1$ labels with confidence scores $\epsilon$ and $1-\epsilon$ for some small value $\epsilon$ (set to $0.01$ in our experiments).

\begin{figure}[ptb]
\centering
    \includegraphics[trim={0.6cm 0.5cm 0.55cm 0.55cm}, clip,width=\gphsize\columnwidth]{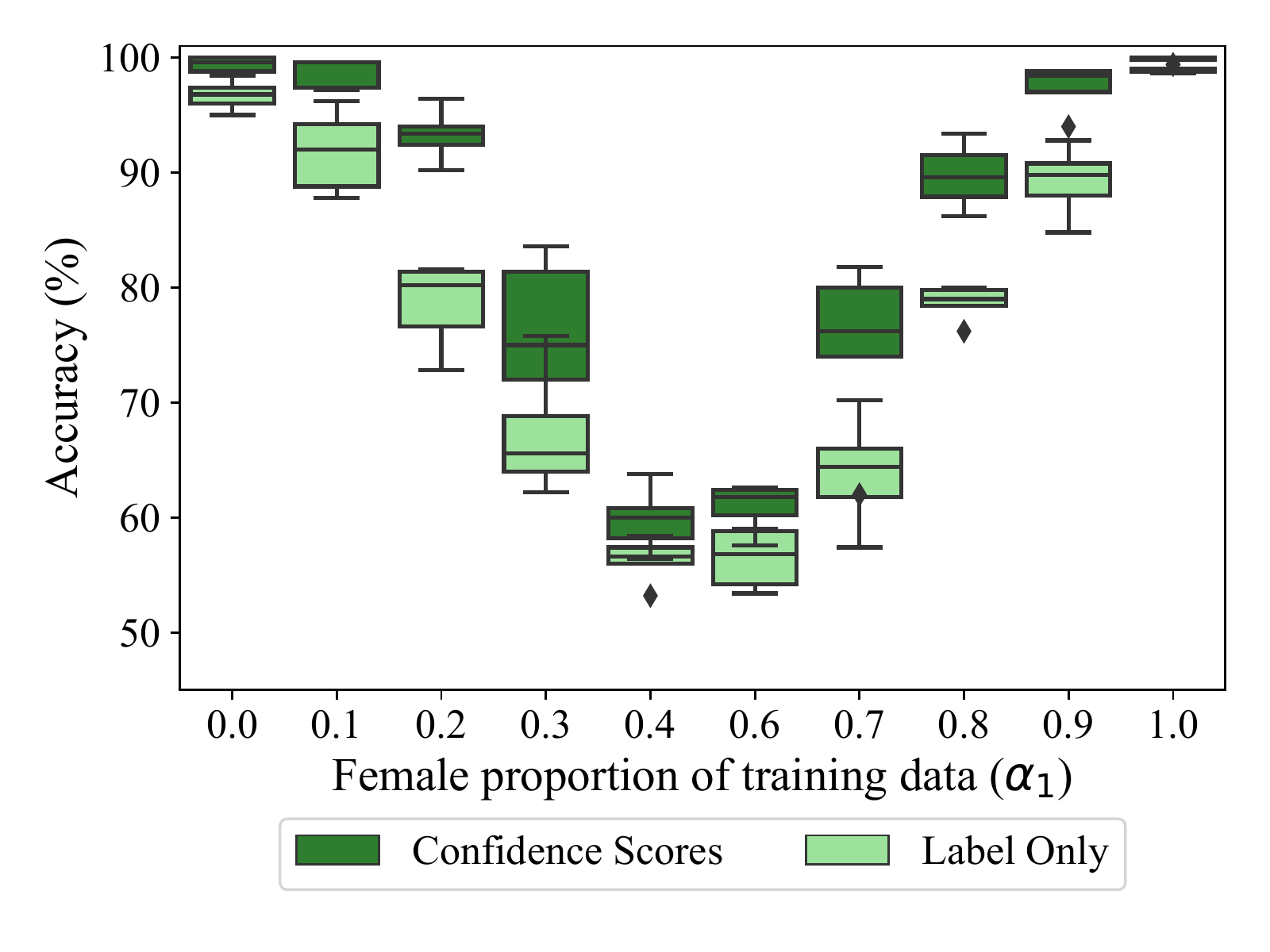}
\caption{Comparing the distinguishing accuracy for the \kldtest\ for \celeba\ (Sex), when the target model returns prediction confidence scores and when it returns only prediction labels. Performance drops most for certain ratios like 0.2 and 0.8, but remains high and roughly the same for more extreme ratios like 0.0, 0.1, and 1.0.} 
\label{fig:celeba_kl_full_vs_label}
\end{figure}

We observe mixed trends across datasets and attacks. For instance, switching to the label-only setting has little impact in the case of \censusn, while mean distinguishing accuracies drop by more than 8\% (\neff\ drops by more than half) for \celeba\ (\autoref{tab:label_only}). However, the drop in performance for \celeba\ is not uniform across all ratios. Inference risk is still quite high for many values of $\alpha$ (\autoref{fig:celeba_kl_full_vs_label}). Similar trends hold for \boneage, where distinguishing accuracy is $>75\%$ for all ratios. We also experiment with using probabilistic sampling to extract more information.
For each datapoint, we sample $k$ random points in its neighborhood by adding random noise from $\mathcal{N}(0, \sigma^2)$ to each feature, and average the generated label predictions to estimate confidence scores. We observe slight improvements in attack performance from the sampling, at the cost of additional queries.

\begin{table}[tb]
\centering
\scalebox{0.94}{
\begin{tabular}{lccc}
 \toprule
\multirow{2}{*}{\bf Dataset/Task} & \multirow{2}{*}{\bf Confidence Scores} & \multicolumn{2}{c}{\bf Prediction Label} \\
& & Direct & Sampling \\
 \midrule
 \censusn\ (Sex) & 82.5 (4.2) & 77.3 (3.3) & 80.5 (3.7)\\
 \celeba\ (Sex) & 85.3 (3.2) & 77.8 (1.4) & 79.3 (1.6) \\
 \boneage\ (Age) & 99.8 (22.6) & 96.3 (12.7) & 97.1 (13.3)\\
 \bottomrule
\end{tabular}
}
\caption{Effectiveness of label-only attacks. Each value is the observed mean distinguishing accuracy (\%) (with mean \neff\ in parentheses) for \kldabbr\ (as computed in Table~\ref{table:datasets_desc}). The label-only setting leaks less information, but the attacks still are effective even when confidence scores are unavailable. `Direct' uses a single query, while `Sampling' uses 10 samples around each test point.}
\label{tab:label_only}
\end{table}

\section{Defenses} \label{sec:defenses}

Several defenses against distribution inference have been proposed, but most of them (except differential privacy~\cite{ganju2018property}, which has shortcomings as we discuss later) have not been actually evaluated. Like most defenses designed to limit privacy leakage, these defenses involve adding noise in some parts of the training process. This can include the data itself~\cite{ateniese2015hacking} or model parameters~\cite{ganju2018property,melis2019exploiting}.
One notable exception is work by Hartmann et. al.~\cite{Hartmann2022DistributionIR}, where the authors study causes of leakage in distribution inference attacks, and evaluate mitigation strategies based on causal learning (IRM~\cite{arjovsky2019invariant}), correcting inductive biases, and increasing the amount of training data, for synthetic datasets.
We evaluate some of these noise-based defenses in Section~\ref{sec:noise_defenses}, and find that they seem unlikely to successfully mitigate distribution inference risks. Our exploration of inference risk with model generalization reveals interesting trends and a potential trade-off between learning and inference risk 
(Section~\ref{sec:generalization}). Section~\ref{sec:rebalance} introduces and evaluates a simple defense based on data re-sampling, which can prevent distribution inference in settings where the model trainer knows which distributional property to hide.

\shortsection{Prior Work}
Unlike membership inference where differentially private training can provide a guaranteed bound on inference risk, there are no defenses against distribution inference from trained models with theoretical guarantees. Chen and Ohrimenko~\cite{chen2022} recently proposed a defense mechanism that builds upon formal notions of distributional privacy~\cite{zhang2022attribute} to protect against property inference attacks on statistical queries. This is the first known theoretically-grounded defense against distribution inference, but it does not apply to protecting machine-learning models.

The only previous defense that has demonstrated meaningful protection against distribution inference attacks on machine learning models is NoSnoop~\cite{ma2021nosnoop}, proposed for a collaborative learning setting. In their threat model, the adversary seeks to infer sensitive information about exact training batches and has access to intermediate model losses from clients. The defense works utilize a discriminator-generator setup, where gradient updates are used to minimize property leakage while preserving task-based utility.
Although this defense is highly effective, it defends against a very narrow type of configuration, including properties limited to the presence/absence of sensitive data.

Recent work by Stock et al.~\cite{stock2022property}  proposes a defense against distribution inference based on gradient updates from meta-classifiers. The victim, being aware of two distributions that an adversary may test for, trains multiple copies of its models with these training distributions. Then, it trains a meta-classifier and computes gradient updates for its own local models such that inference risk is minimized.
This method requires the victim to train hundreds and thousands of models locally for the meta-classifier, leads to large drops in task performance, and does not generalize to settings where the victim and adversary use different kinds of meta-classifiers.

Other proposed defenses include removing sensitive attributes from features~\cite{zhang2021leakage}, using node-multiplicative transforms, or encoding arbitrary information into the models~\cite{ganju2018property}. Since black-box attacks only utilize relationships between inputs and model outputs, they are unaffected by such changes as long as model functionality remains unaffected. Additionally, the Permutation-Invariant Network architecture can be modified to have some form of scale-invariance as well, which in turn can also bypass such multiplicative defenses. Further, these defenses seem unlikely to diminish black-box attacks, which our experiments have shown to be more effective than known white-box attacks, hence we do not evaluate them here.

\subsection{Noise-Based Defenses} \label{sec:noise_defenses}

Several proposed defenses against distribution inference involve adding noise in various ways---differentially privacy training incorporates crafted noise in the training process and label poisoning adds noise to the training data. We also consider using adversarial training, which augments training with adversarial perturbations.

\shortsection{Differentially Private Training} \label{sec:defense_dp}
Differential privacy (DP) is a formal privacy notion that provides theoretical guarantees that bound an adversary's ability to distinguish between neighboring input datasets from the output of a computation. Differential privacy can provide theoretical bounds limiting membership inference.
Evaluations by Ateniese et al.~\cite{ateniese2015hacking} suggest differentially private training is not an effective defense against distribution inference attacks. However, their experiments used a setup with some overlap between the victim's and adversary's data, so it is possible the observed lack of protection is related to the overlapping data available to the adversary. Although differential privacy in itself does not guarantee protection against distribution inference, evaluating risk for models trained with these guarantees can help better understand how such noise-based mechanisms can affect inference risk, and assess the vulnerability of models meant to provide membership inference privacy. Empirical evidence can thus be beneficial and more concrete than relying on pure intuition (or argumentative reasoning about why a defense may or may not work).

\begin{figure}[tb]
\centering
    \includegraphics[trim={0.6cm 0.5cm 0.55cm 0.55cm}, clip,width=\gphsize\columnwidth]{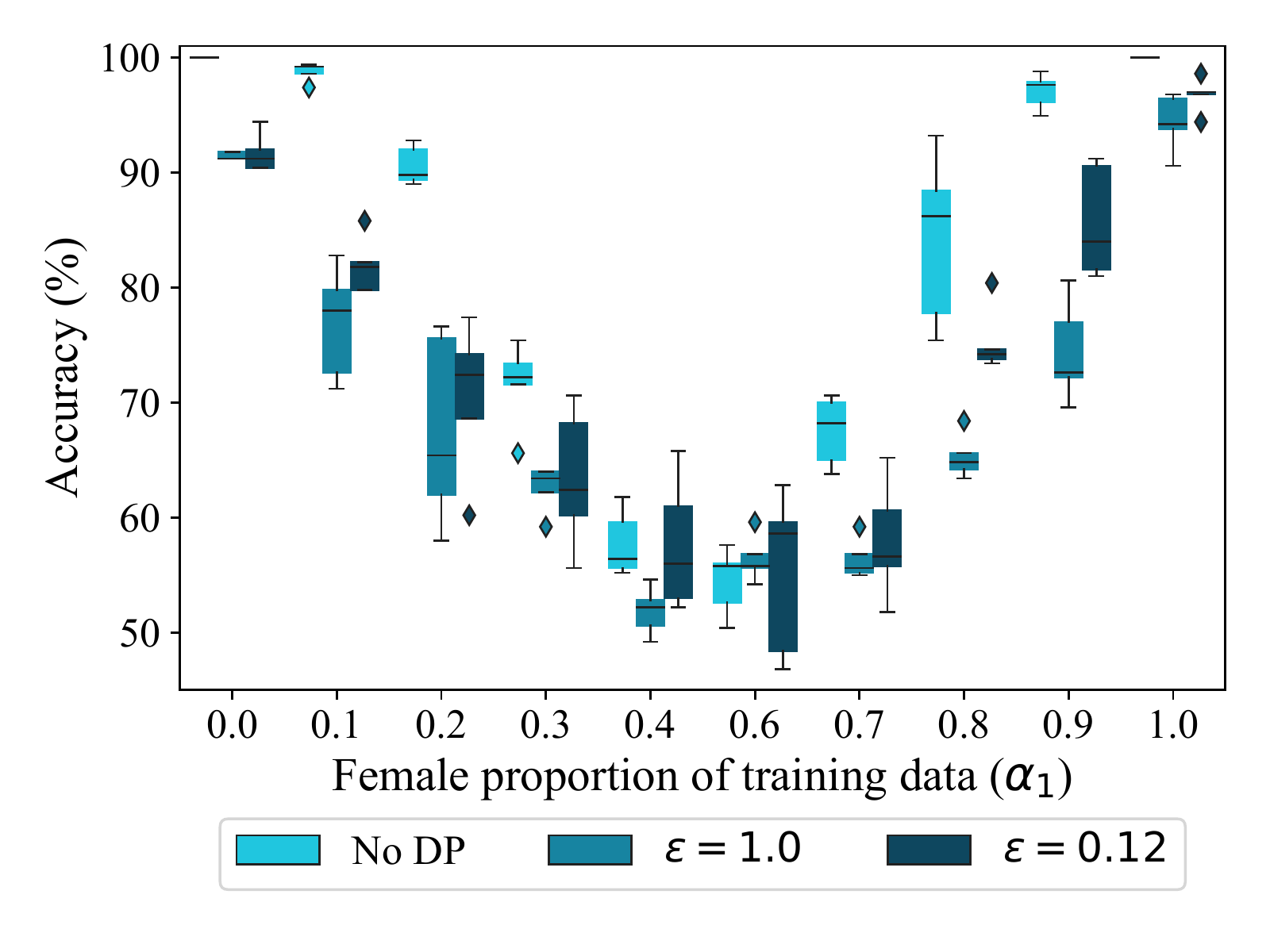}
\caption{Distinguishing accuracy for different for \censusn\ (Sex), using \kldtest. Attack accuracy drops with stronger DP guarantees (decreasing privacy budget $\epsilon$).}
\label{fig:dp_bb}
\end{figure}

We use DP-SGD~\cite{abadi2016deep} to train victim models with Re\'nyi Differential Privacy~\cite{mironov2017renyi}, with privacy loss budgets of $\epsilon = 1.0$ and $\epsilon = 0.12$, with $\delta = 4.9 \times 10^{-6}$. We evaluate this defense on \censusn, since it is the only tabular dataset with non-trivial inference leakage. We observe a drop in distinguishing accuracies, but inference risk stays high for ratios further away from $\alpha_0=0.5$ (Figure~\ref{fig:dp_bb}). 

We hypothesize that this drop in effectiveness may not be because of the differential privacy noise itself, though, but could be because either the model does not learn the distribution well enough (and hence does not reveal it), or it produces arbitrary differences that cause a mismatch between the victim's models trained with using DP-SGD and the adversary's shadow models trained without privacy noise. Inspection of task accuracy for the differential-privacy models suggests lower learning effectiveness as one potential factor (\autoref{tab:defenses}). To test whether the decrease in prediction accuracy is mainly due to arbitrary differences in the models, we evaluate results for the setting where the adversary also trains its models using DP-SGD with the same privacy loss budget. Compared to an adversary that does not use DP, there is a clear increase in inference risk---mean distinguishing accuracy increases to 86.4\% (\neff=2.9) for $\epsilon=1.0$, and 91.5\% (\neff=4.8) for $\epsilon=0.12$ (compared to 82.5\%, i.e. \neff=4.2 without any DP).

Assuming adversary's knowledge of the use of differentially-private training and the specific privacy loss budget is not a far-fetched assumption. Organizations that release differentially private models often document their exact levels of privacy budget~\cite{thakurta2017learning, abowd20222020}. An adversary in such scenarios can thus train its models with the same privacy parameters.

\begin{figure}[tb]
\centering
    \includegraphics[trim={0.6cm 0.5cm 0.55cm 0.55cm}, clip,width=\gphsize\columnwidth]{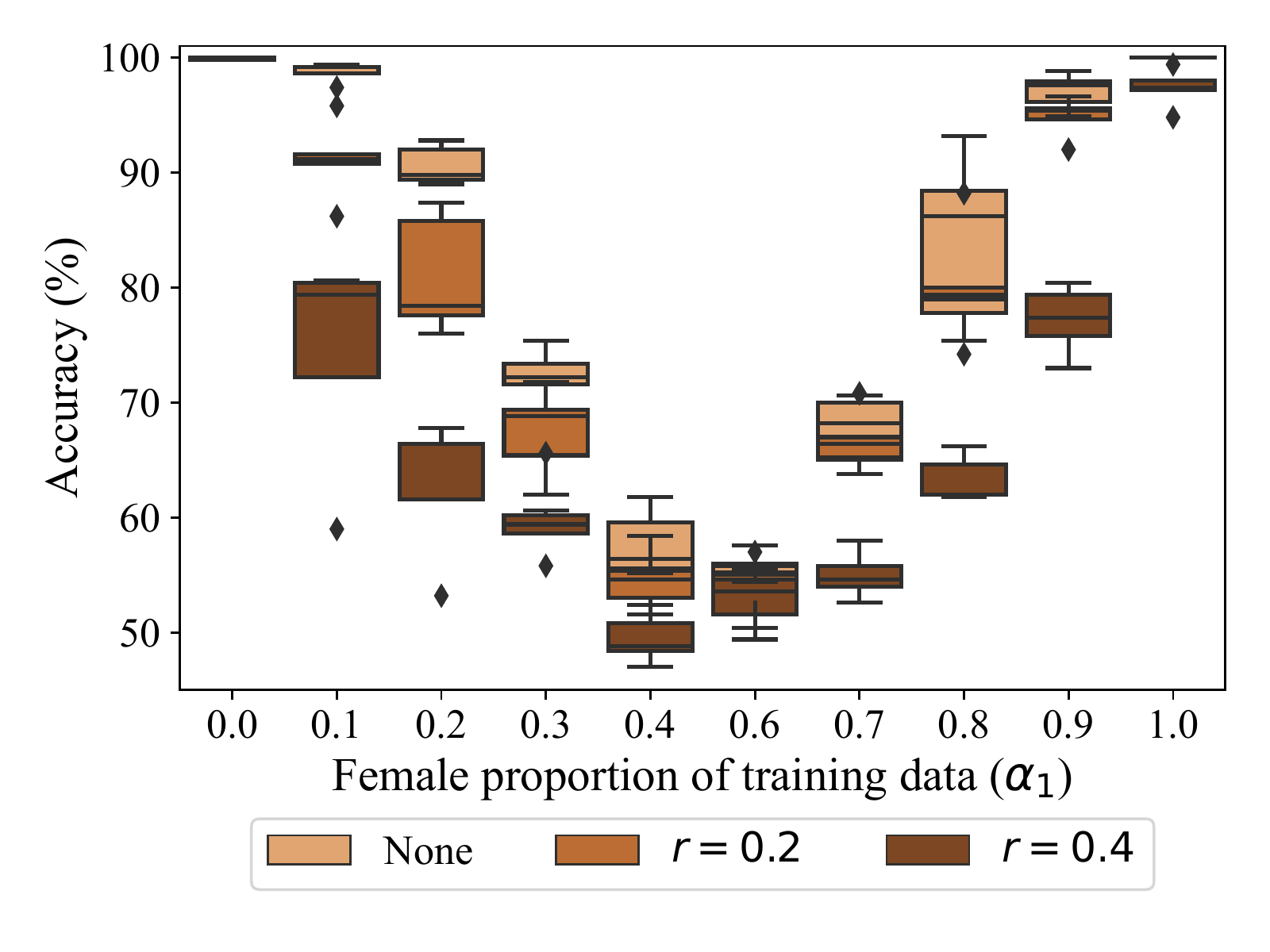}
\caption{Distinguishing accuracy for different for \censusn\ (Sex), for varying levels of label poisoning. Inference risk drops considerably with increasing levels of label poisoning, but is also followed with non-trivial drops in task accuracies.}
\label{fig:census_label_noise}
\end{figure}

\shortsection{Label Poisoning} \label{sec:defense_label_poison}
Ganju et al.~\cite{ganju2018property} proposed to mitigate distribution inference by adding noise to the training data via label poisoning. The underlying idea is to perturb the training data in a way that will alter the model parameters and make the adversary's task harder. Although changing data labels is prone to detrimentally harming the model's task performance, a model trainer may be able to find an acceptable trade-off between accuracy and inference risk.
For a given noise ratio $r$, the defense comprises randomly flipping task labels for $r$ fraction of the training data. We evaluate this defense for \celeba\ (Male) and \boneage\ (Age) with a label noise ratio of 0.2, and \censusn\ (Gender) for label noise ratios 0.2 and 0.4. As expected, this defense harms task performance (\autoref{tab:defenses}), reducing task accuracy: by $\sim1-2\%$ for $r=0.2$ for all three datasets, and $\sim3\%$ for $r=0.4$ off \censusn.
Average inference risk drops for \censusn, but remains is still quite high for ratios like $\alpha_1<0.2$ and $\alpha_1>0.8$, as shown in Figure~\ref{fig:census_label_noise}.
It may be possible to find a desirable tradeoff for a simple task like \censusn, but this approach is not effective for more complex tasks. For instance, using a label noise ratio of $0.4$ in \celeba\ completely destroys task performance, reducing the classifier to only slightly better than random guessing.

\begin{figure}[ptb]
\centering
    \includegraphics[trim={0.6cm 0.5cm 0.55cm 0.55cm}, clip,width=\gphsize\columnwidth]{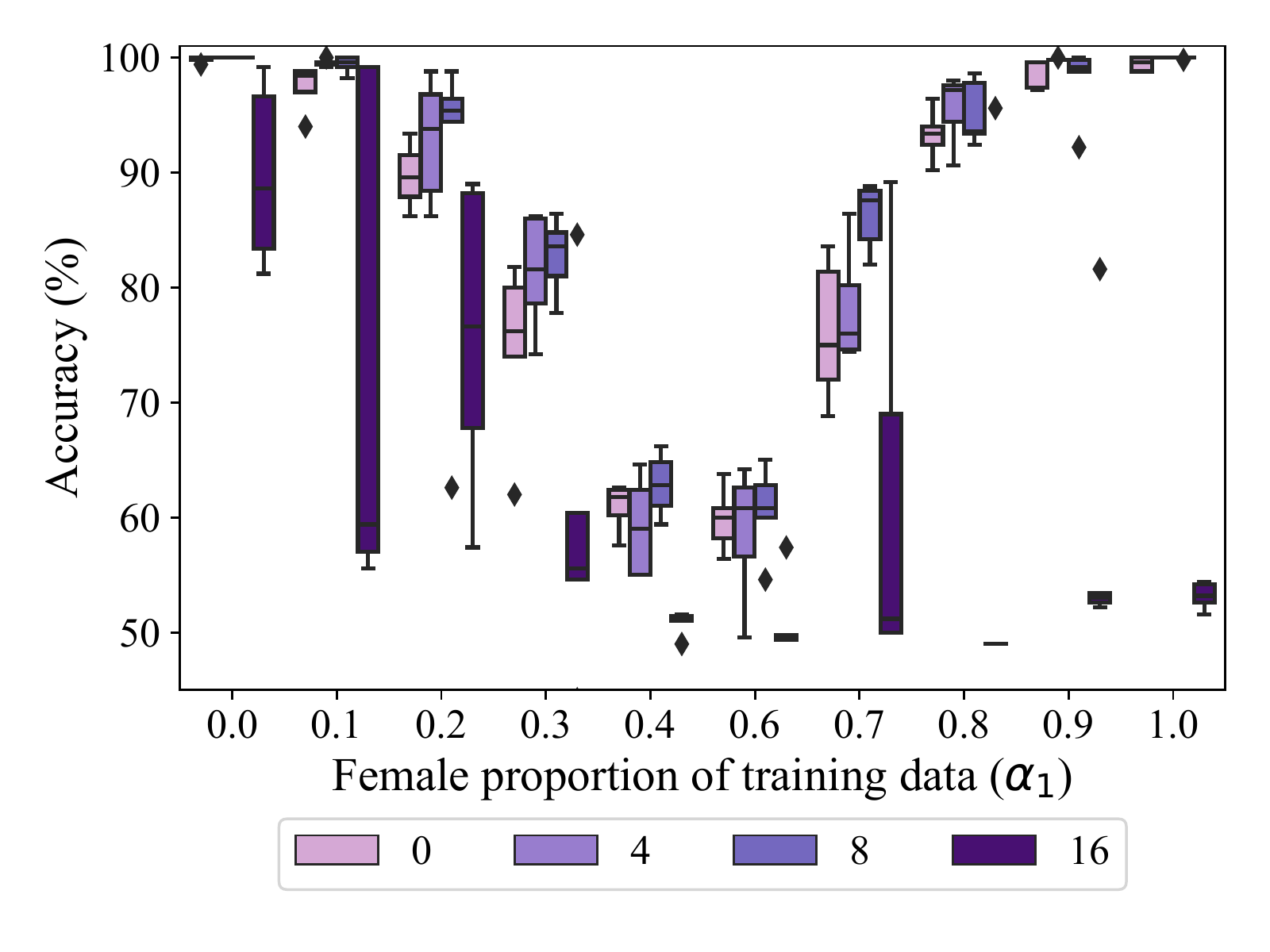}
\caption{Distinguishing accuracy for different using \kldabbr, for varying levels of adversarial robustness $\epsilon$ (/255) in $L_\infty$ norm, for \celeba\ (Sex). Inference risk lowers with increasing robustness.}
\label{fig:celeba_robust_male}
\end{figure}

\shortsection{Adversarial Training} \label{sec:adv_robustness}
Adversarial training~\cite{madry2018towards} involves using a training loss function that encourages the model to learn features that are robust to perturbations in the input, and produces models that are less prone to overfitting dataset-specific patterns~\cite{ilyas2019adversarial}. This can be especially useful when the data includes properties that are not correlated with the  task, and a model should not capture signals related to the irrelevant property. A model trained with adversarial robustness objective, using this reasoning, should be less susceptible to distribution inference. To test this hypothesis
and explore the impact of training for robustness. We train adversarially-robust models for varying $L_\infty$ norms for the Gender and Age properties on \celeba, since the other datasets are either tabular or do not contain sufficient samples for adversarial training with acceptable performance. \autoref{fig:celeba_robust_male} shows distinguishing accuracies for varying settings for the perturbation strength used in adversarial training.
Training for adversarial robustness, as documented in the literature, leads to drops in task accuracy.

We observe very interesting trends with respect to inference risk. Risk increases with increasing perturbation strength ($\epsilon$) until $8/255$, and then drops to near-zero (\autoref{fig:celeba_robust_male}).

\begin{table}[tb]
\centering
\begin{tabular}{lcccc}
 \toprule
 \multirow{2}{*}{\bf Dataset/Task} & \multicolumn{4}{c}{\bf Adversarial Training ($\epsilon$)}\\
 & 0/255 & 4/255 & 8/255 & 16/255\\
 \midrule
 \celeba\ (Sex) & 85.3 (3.2) & 86.8 (5.3) & 88.3 (5.2) & 58.9 (0.2)\\
 \celeba\ (Age) & 86.3 (2.5) & 90.0 (5.5) & 88.9 (6.4) & 83.6 (2.0) \\
 \bottomrule
\end{tabular}
\caption{Impact of adversarial training. Values are mean distinguishing accuracies (\%) (with mean \neff; as computed in \autoref{table:datasets_desc}) for \kldabbr\ on models trained with adversarial robustness, with varying norms $\epsilon$ (/255) of perturbation budget ($L_\infty$ norm).}
\label{tab:robustness}
\end{table}
Since adversarial training helps models remove focus from spurious correlations, it is naturally aligned with causal inference~\cite{zhang2022adversarial}. This then leads to these models using signals relevant to the property being inferred (like Age or Sex) even more, since it is related to the task. However, as this perturbation norm increases, task accuracy drops accordingly, thus leading to lower inference risk since the model itself performs poorly at learning causal connections, like the one between the property being inferred and the given task. One notable exception here is $\epsilon=8/255$ for \celeba\ (Age), where inference risk seems to slightly increase. One possible explanation is the stronger age (property) and sex (task) relationship in this case, leading to the causal relationship-accuracy tradeoff leaning in favor of the former, in terms of inference risk.

\subsection{Generalization}
\label{sec:generalization}

The defenses discussed in Section~\ref{sec:noise_defenses} have one thing in common: they nearly always lead to non-trivial drops in task performance. This is not only unacceptable for most deployments, but raises the question of whether the defenses are doing anything useful or just reducing distribution inference by producing models that learn the underlying distribution less well. Concretely, we observe a positive correlation between model task accuracy and mean \neff\ values (Pearson's correlation coefficient $>0.55$). 
A model with poor task performance possibly fails to learn useful signals from the training distribution, and is thus cannot leak properties it has not learned; while good performance means a model has learned the distribution well and is prone to more leakage.

\begin{figure}[tb]
\centering
    \includegraphics[trim={0.5cm 0.6cm 0.4cm 0.6cm}, clip,width=\gphsize\columnwidth]{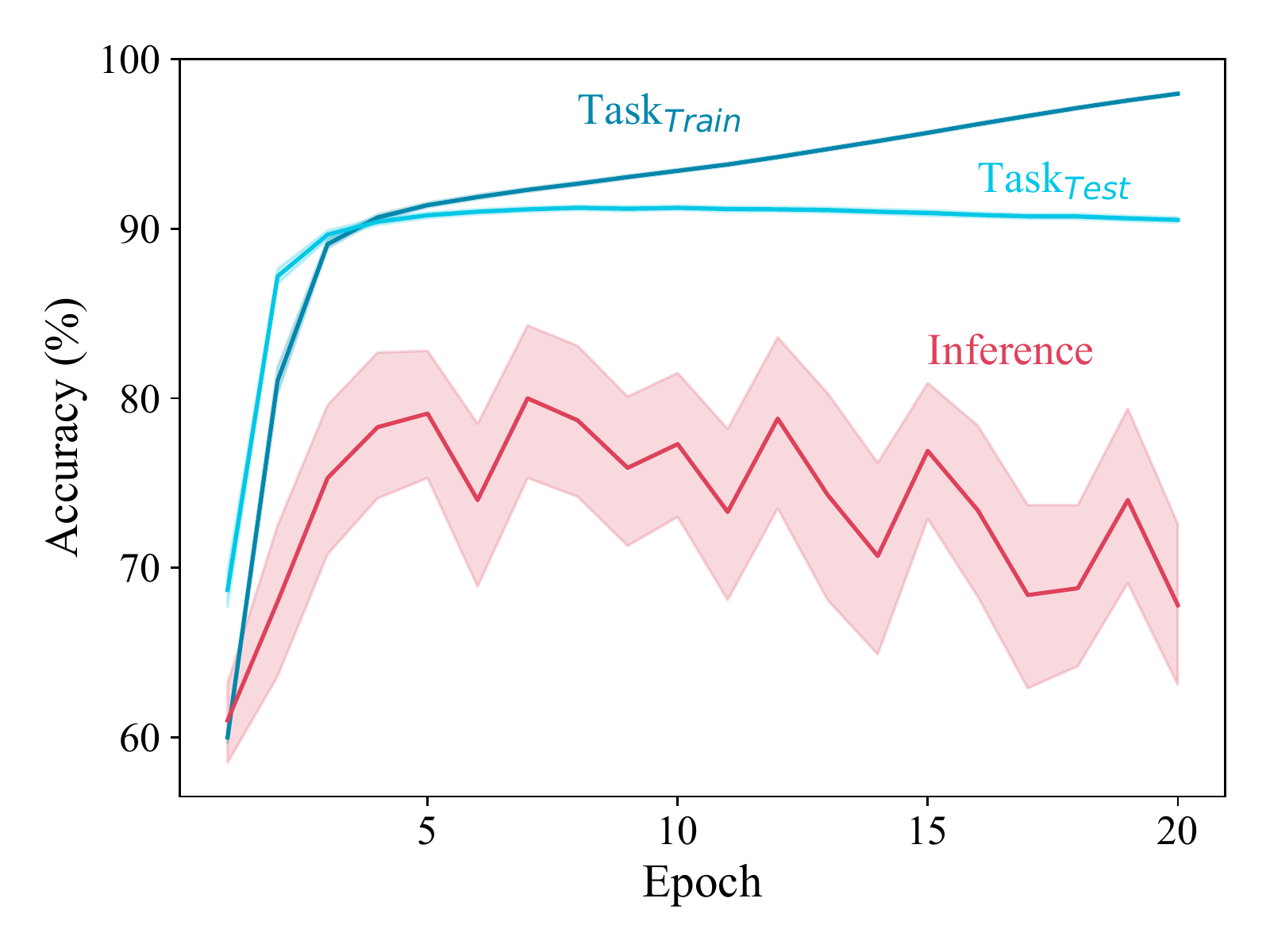}
\caption{Mean distinguishing accuracy (as computed in \autoref{table:datasets_desc}) of the \kldtest\ on \celeba\ (Sex), for varying number of training epochs for victim models. Shaded regions correspond to error bars. Distribution inference risk increases as the model trains, and then starts to decrease as the model starts to overfit.}
\label{fig:generalization_trend_census}
\end{figure}

To investigate these correlations, we inspect trends between inference risk and generalization across training epochs. For this experiment, we train the models longer than the other experiments (which were optimally selected for best generalization using validation data), allowing us to better study trends between overfitting and inference risk.
We observe interesting trends in inference risk with model training. In most cases, inference risk is high after even one epoch of model training (\autoref{fig:generalization_trend_census}). This is especially surprising because the model takes a few epochs to get good performance on the task itself, but shows that the model is learning and exposing aspects of the distribution even early in its training.

These trends clearly suggest that model under-training is not a feasible defense. Training beyond minimum generalization gap does lead to significantly reduced distribution inference risk. However, this region of the training corresponds to overfitting, which is known to be positively correlated with increased risk to membership inference~\cite{yeom2018privacy}. Thus, a model trainer that is willing to overfit its models to avoid distribution inference adversaries would risk making the model more vulnerable to membership inference.

\subsection{Re-Sampling Data} \label{sec:rebalance}

If the victim is aware of the property that an adversary might target, or only has a few known properties of the distribution that it wants to protect, the easiest mitigation is to just modify the training distribution (or the sampling mechanism) such that the property is no longer present for the training dataset. Knowing the particular property to hide is a plausible assumption that is often assumed in work on distribution inference defenses. For instance, Chen and Ohrimenko~\cite{chen2022} propose a theoretically-grounded defense that builds upon the distributional privacy framework~\cite{kawamoto2019local} and modifies feature values to provide privacy guarantees against distribution inference adversaries. 

Zhou et al.\ \cite{zhou2021property} propose over-sampling to reduce inference risk, but do so by adding new samples to their training data. Although this defense eliminates distribution risk (at the cost of model performance), the availability of new data is not always possible. Model trainers typically use all available data.

We explore two variations of re-sampling defenses: \emph{over-sampling} and \emph{under-sampling}. In both cases, the model trainer re-samples data from its available datasets such that the resulting dataset is indistinguishable from a dataset sampled from a different distribution. For over-sampling we experiment with two flavors: using simple replacement and over-sampling based on inserting augmented data. 
These defenses rely on the key assumption that the model trainer knows the property they want to hide, and that there are a only few such properties so re-sampling to hide the desired properties will not unduly hard the model's task performance. When this assumption holds, resampling defenses can virtually eliminate inference risk. 
We evaluate this defense on configurations with low (\celeba--Sex), medium (\censusn--Sex), and high (\boneage--Age) inference risk to measure the impact of this defense.

\shortsection{Under-Sampling}
The model trainer can simply under-sample its data such that the resulting dataset has a ratio corresponding to some other distribution. For example, a model trainer, with a dataset containing $70\%$ females who wants to hide the ratio of females in the dataset from an inference adversary can simply under-sample examples with the `female' attribute such that its data is balanced. This defense should prevent any disclosure about the pre-sampled distribution since there should be no difference between the cases where the training data was balanced to begin with and when it was adjusted with this defense, so long as the distribution is not distorted by the under-sampling. In our experiments, we find that under-sampling lowers inference risk significantly, but does not completely eliminate it---mean distinguishing accuracy drops below $54\%$ (\neff$<$0.1) for \celeba\, with significantly lower leakage for \censusn\ ($<57\%$, i.e. \neff$<$0.1) and \boneage\ ($\sim60\%$, i.e. \neff=0.3) (Table~\ref{tab:defenses}).

\shortsection{Over-Sampling}
A model trainer not willing to sacrifice available training data by under-sampling may prefer to over-sample. The most basic variant over-samples the data before training begins, duplicating training records, and then trains its models like usual. Although this defense leads to the complete utilization of data, the presence of repeated data can be revealing and may reveal the property the adversary wants to hide to an adversary aware of the defense. It could, for instance, lead to a change in group-wise accuracies, which an adversary can learn to identify and still succeed at distribution inference. 

\shortsection{Augmentation Based Over-Sampling}
The ideal scenario for the defense would comprise of injecting fresh labeled data to adjust the desired property, as was assumed by Zhou et al.~\cite{zhou2021property}. However, labeled data is scarce and may be expensive to acquire, and using techniques like pseudo-labeling can still leak information. In such scenarios, the model trainer can use augmentation techniques to synthetically generate additional samples. This avoids repeating samples, and may have the added benefit of potentially increasing the model's robustness to augmentations. But, the use of augmented data in an imbalanced way may still reveal information to a distribution inference adversary. For this defense, we focus only on the \celeba\ dataset, since designing augmentation for tabular datasets is much harder, and augmentations for \boneage\ are limited. Task accuracy remains comparable and inference risk drops significantly (slightly higher than other forms of sampling), but is not completely eliminated and still higher than standard under and over-sampling.

\shortsection{Impact on Fairness}
This form of re-sampling is common in research related to improving fairness in machine learning~\cite{mehrabi2021survey}, commonly known as ``unbiasing". However, re-sampling data can impact different sub-groups and populations of the distributions unequally, creating issues related to fairness in model predictions~\cite{krasanakis2018adaptive}. To investigate such potential impacts, we measure the impact of under-sampling and over-sampling-based mitigation strategies on fairness. We compare the precision and recall for another group and its possible values, for both undersampling and oversampling.
Re-sampling based defenses have negligible impact on fairness in the case of \celeba, but result in disparate impacts of both under/over-sampling on the two groups. for \censusn\  (Table~\ref{tab:fairness_impact_censusn}). For instance, over-sampling from a ratio $\alpha<0.5$ lowers both precision and recall for whites, but increases recall and decreases precision more greatly for not-whites. These changes are even more severe for \boneage, where changes in precision can be as high as $20\%$ in opposite directions for different groups.

\begin{table}[tb]
\centering
\renewcommand{\arraystretch}{1.1}
\begin{tabular}{lccccc}
 \toprule
 \multirow{2}{*}{\bf Re-Sampling} & \multirow{2}{*}{\bf $\alpha$} & \multicolumn{2}{c}{\bf Precision} & \multicolumn{2}{c}{\bf Recall} \\
 & & not-white & white & not-white & white \\
\midrule
 \multirow{2}{*}{Under-Sampling} & $< 0.5$ & {$\downarrow2\%$} & {$\sim$} & {$\uparrow1\%$} & {$\downarrow1\%$}\\
 & $>0.5$ & {$\downarrow1\%$} & {$\sim$} & {$\downarrow1\%$} & {$\sim$}
 \\[1ex]
 \multirow{2}{*}{Over-Sampling} & $< 0.5$ & {$\downarrow2\%$} & {$\downarrow1\%$} & {$\uparrow1\%$} & {$\downarrow1\%$} \\
 & $>0.5$ & {$\downarrow1\%$} & {$\downarrow1\%$} & {$\downarrow1\%$} & {$\downarrow1\%$} \\
 \bottomrule
\end{tabular}
\caption{Relative change (\%) in precision and recall metrics for white and not-white (race attribute), for \censusn\ (gender) for under-sampling and over-sampling. We consider cases where data for males ($\alpha<0.5$) or females ($\alpha>0.5$) is under-sampled for equalization.}
\label{tab:fairness_impact_censusn}
\end{table}

\renewcommand{\thefootnote}{\fnsymbol{footnote}}

\begin{table*}[tb]
\centering
\begin{tabular}{lccc}
 \toprule
 \multirow{2}{*}{\bf Defense} & &  \multicolumn{2}{c}{\bf Distinguishing Accuracy (\neff)} \\
  & \bf  Task Accuracy (\%) & $\alpha_1=0.2$ & Mean \\
 \midrule
 \multicolumn{4}{c}{\censusn\ (Sex)} \\[1ex]
 No Defense & {$77.9\pm0.9$} & 89.8 (2.1) & 82.5$\pm$17.9 (4.2$\pm$5.3)\\
 DP ($\epsilon=1.0$) & {$77.0\pm1.0$} & 65.4 (0.2) & 69.3$\pm$14.6 (0.6$\pm$0.7)\\
 DP ({$\epsilon=0.12$}) & {$75.6\pm1.0$} & 72.4 (0.5) & 73.4$\pm$14.8 (0.8$\pm$0.9)\\
 Label Poisoning ($r=0.2$) & {$77.3\pm1.0$} & 78.4 (0.8) & 78.9$\pm$17.4 (3.5$\pm$5.4)\\
 Label Poisoning ($r=0.4$) & {$74.9\pm1.2$} & 66.4 (0.2) & 70.0$\pm$17.9 (1.9$\pm$4.2)\\
 Under-sampling & {$77.5\pm0.5$} & 50.0 ($<$0.1) & 56.7$\pm$6.8 (0.1$\pm$0.1)\\
 Over-sampling & {$77.3\pm0.6$} & 50.0 ($<$0.1) & 51.9$\pm$2.5 ($<$0.1$\pm$0)\\
 \midrule
 \multicolumn{4}{c}{\boneage\ (Age)} \\[1ex]
 No Defense & {$65.8\pm2.0$} & 99.9 (20.1) & 99.8$\pm$0.4 (22.6$\pm$4.2)\\
 Label Poisoning ($r=0.2$) & {$64.3\pm2.3$} & 99.9 (20.1) & 95.7$\pm$6.2 (12.1$\pm$7.1)\\
 Under-sampling & {$65.4\pm3.2$} & 73.4 (0.5) & 59.1$\pm$13.3 (0.3$\pm$0.5)\\
 Over-sampling & {$64.6\pm2.8$} & 70.4 (0.4) & 60.2$\pm$11.2(0.3$\pm$0.4)\\
 \midrule
 \multicolumn{4}{c}{\celeba\ (Sex)} \\[1ex]
 No Defense & {$91.6\pm0.8$} & 89.6 (2.1) & 85.3$\pm$15.8 (3.2$\pm$2.7) \\
 Label Poisoning ($r=0.2$) & $90.0 \pm 5.0$ & 82.0 (1.1) & 78.3$\pm$15.6 (1.2$\pm$1.0)\\
 Adv. Training ($\epsilon=4/255$) & $90.4\pm0.8$ & 93.8 (3.1) & 86.8$\pm$16.4 (5.3$\pm$5.2)\\
 Adv. Training ($\epsilon=8/255$) & $88.5\pm1.2$ & 95.4 (3.7) & 88.3$\pm$15.0 (5.2$\pm$4.9)\\
 Adv. Training ($\epsilon=16/255$) & $75.7\pm11.9$ & 76.6 (0.7) & 58.9$\pm$13.1 (0.2$\pm$0.4)\\
 Under-sampling & {$90.8\pm1.1$} & 50.0 ($<$0.1) & 53.7$\pm$6.1 ($<$0.1$\pm$0.1)\\
 Over-sampling & {$90.6\pm0.8$} & 50.0 ($<$0.1) & 53.8$\pm$4.1 ($<$0.1$\pm$0.1)\\
 Augmentation-based over-sampling& {{$91.7\pm1.6$}} & {{74.8 (0.6)}} & {{61.0$\pm$14.5 (0.3$\pm$0.5)}} \\
 \bottomrule
\end{tabular}
\caption{Effectiveness of considered defenses. Each distinguishing accuracy (and corresponding \neff) reported is the observed leakage of \kldabbr. The first results are for predicting between $\alpha_0=0.5$ and $\alpha_1 = 0.2$; the last column reports mean distinguishing accuracy (with mean \neff\ in parentheses) as described in \autoref{table:datasets_desc}.  Mean distinguishing accuracies (and \neff\ numbers) are reported with $\pm$ standard deviation, over different $\alpha_1$ values.
Most noise-based defenses harm model task accuracies, and the only defenses that diminish leakage without harming task accuracy are based on data re-sampling.
}
\label{tab:defenses}
\end{table*}

\shortsection{Adaptive Attacks}
An adversary with knowledge of the under-sampling approach may be able to derive the original distribution by estimating the size of the training data to learn the sampling ratio. The strongest adversary would be one that starts with knowledge of specific records in the original training dataset, and can use membership inference attacks to estimate how many of those records are included in the under-sampled dataset. We evaluate such attacks in  Appendix~\ref{app:adaptive_attacks}, and find they are unlikely to be effective without dramatic improvements to membership inference attacks.
\section{Limitations and Conclusions}
\label{sec:conclusion}

Distribution inference attacks are known to reveal sensitive properties about underlying training distributions, but their effectiveness has been established only in very controlled settings so far. Our work advances understanding of this risk in more realistic settings, but is still far from understanding all of the issues that might impact a real deployment and attack.

Our proposed black-box attacks are highly efficient and maintain their effectiveness even when access to exact prediction probabilities is unavailable. Our experiments find that differences in model architectures harm inference accuracies, but the impact is not very severe as long as models with similar complexity are used.
Even the lack of common feature extractors, a common setting in many evaluations in the literature, does not completely eliminate inference risk.

Like nearly all inference privacy work, we assume an adversary with access to a dataset that matches the underlying distribution (in this case, before the transformation to the actual training distribution as modeled by \gzero\ and \gone). This is a strong assumption, which may be realistic in some cases but is often unlikely. All our attacks (and nearly all previous ones) require representative data for training models locally.  
Exploring adversaries with limited data access to these distributions and how it impacts inference risk is left as part of future work.

Configurations beyond single-party learning have seen growing interest lately. Active adversaries with data poisoning capabilities have been demonstrated to be highly effective in both single-party~\cite{mahloujifar2022property} and multi-party~\cite{wang2022poisoning} settings. Our proposed attacks are highly potent for many configurations, but the exact impact of different training setups like federated learning remains largely unexplored.

The general approach to achieve security and privacy for machine-learning models is to add noise, but our evaluations suggest this approach is not a principled or effective defense against distribution inference. The main reductions in inference accuracy that result from these defenses seem to be due to the way they disrupt the model from learning the distribution well, so observed reductions in inference risk are related to drops in task performance. Our experiments with different model architectures and differentially private training support this---inference risk increases significantly when the victim and adversary use the same learning algorithms or model architectures. The only reliably effective defense from our experiments is to re-sample data, which depends on the assumption that the model training is aware of the adversary's inference goals (or at least of the properties that should be protected). These re-sampling defenses too are not perfect, however, as they seem to negatively impact fairness of groups related to the property attribute.

There is a need for more theoretical connections between distribution inference risk and general useful notions of machine learning, like model complexity and fairness. Our work suggests such connections do exist, and we hope they will be better understood as both empirical and theoretical understanding of inference privacy advances.



\subsection*{Availability}
\noindent
Open source code and instructions for reproducing our experiments is available at:\\
\urlstyle{sleo}
\hspace*{2em}\url{https://github.com/iamgroot42/dissecting_distribution_inference}.

\subsection*{Acknowledgements}
\noindent
This work was partially supported by a grant from the National Science Foundation (\#1804603) and support from Lockheed Martin Corporation and Oracle Corporation.

\bibliographystyle{plain} 
\bibliography{references}

\onecolumn
\clearpage
\appendix
\subsection{Experimental Details} \label{app:exp_details}

For each dataset, we create non-overlapping splits of data for the victim and adversary, where the victim has at least double the amount of adversary's data for \graph\ and \boneage, triple for \celeba and \texas, and $4\times$ for \censusn. For each dataset, we simulate $\mathcal{D}$ using the dataset itself. Simulation of distributions with particular $\alpha$ values is is achieved by sampling data with attributes 0 and 1 such that their ratios result in some desired $\alpha$. For properties not based on binary attributes like \graph, this is achieved by pruning nodes iteratively from the graph (while re-computing neighbor counts along the way) to achieve a desired mean node-degree.
We include the datasets and victim/adversary splits used in previous experiments, and include results on two new datasets.
For all of the experiments, we follow the processing pipeline described in Suri and Evans~\cite{suri2022formalizing} to obtain non-overlapping splits. Essentially, both parties sub-sample from their data splits (to achieve specific $\alpha$ values) with different random seeds, and train models on the sampled data.

We perform each experiment five times and report mean values with standard deviation in all of our experiments. For each dataset, we train 250 victim models per distribution. For all black-box attacks, the adversary trains and uses 50 models per distribution for each trial.
For white-box attacks, the adversary trains 800 models per distribution, of which 750 are used for training and 50 as the validation set. For cases with very large models (like DenseNet trained from scratch for \boneage), we use 100 victim models per distribution.

\subsection{Adaptive Attacks Against Under-Sampling}
\label{app:adaptive_attacks}

Assume a more powerful adversary that has access to $m$ training records each corresponding to attribute 0 ($\mathcal{D}_{-}$) and 1 ($\mathcal{D}_{+}$), and the original dataset has size $n$. In this scenario, the adversary is unaware of the original distribution of these attributes ($\alpha$). 
Consider the scenario where the victim utilizes under-sampling on its original distribution as a defense to protect $\alpha$, and re-samples such that both attributes are equally likely.

For our analysis, we assume a near-perfect membership inference adversary, with a false negative rate $\beta$. In such a setup, the adversary can check which of its data (the one with attributes zero, and attributes one) still all tests as members. If all zeros still remain members, then data from ($\mathcal{D}_{+}$) must have been under-sampled, and thus the original $\alpha$ must be $>0.5$.
Assuming that under-sampling is performed by pruning points uniformly at random, the density of members in the resulting data must remain the same. Thus,
\begin{align}
    \frac{m}{\alpha\cdot n} & = \frac{m^{'}}{(1-\alpha)\cdot n} \\  m_{-} & = \beta\cdot m\\
    m_{+} & = m^{'} \cdot \beta
\end{align}
where $m_{+}$ is the number of datapoints (out of the known $m$) with attribute 1 that are inferred as members by the adversary, and $m_{-}$ for attribute 0. $\alpha$ can thus be estimated as $m_{-} / (m_{-} + m_{+})$. By symmetry, the case where original $\alpha<0.5$ yields a similar formula.
We test the risk of this adversary while varying the number of data points $m$, for different values of $\alpha$. The adversary in this case thus directly predicts $\alpha$, and mean square error (MSE) values are computed accordingly for the regression case. We use the R attack from Ye et.al.~\cite{ye2022enhanced}, and use the authors` official implementation, with the FPR set to $0.05$.

\begin{table*}[b]
\centering
\renewcommand{\arraystretch}{1.1}
\begin{tabular}{lccc} 
 \toprule
 \multirow{2}{*}{\bf Dataset/Task} & \multicolumn{3}{c}{\bf Number of known members ($m$)}\\
 & 10 & 100 & 500\\
 \hline
 \censusn\ (Sex) & 9.043 ($<$0.1) & 4.588 (0.2) & 4.078 (0.5)\\
 \boneage\ (Age) & 1.969 (0.1) & 0.486 (0.8) & 0.372 (2.0)\\
 \celeba\ (Sex) & 4.785 ($<$0.1) & 1.466 (0.2) & 1.202 (0.4) \\
 \bottomrule
\end{tabular}
\caption{MSE values (with mean \neff\ in parantheses) for direct regression over $\alpha$ for an adversary that utilizes membership inference to infer $\alpha$ for models trained with under-sampling based defense.}
\label{tab:mi_for_undersampling_regression}
\end{table*}

Similar to the case of binary distinguishing, we use \neff\ to measure the adversary's success. \neff\ for the case of regression is defined in Theorem 4.3 of Suri and Evans~\cite{suri2022formalizing}:
\begin{align}
    \frac{\alpha(1-\alpha)}{\omega}
\end{align}
, for some MSE error $\omega$ while trying to regress over a model corresponding to some training distribution $\alpha$.

MSE values for varying number of members ($m$) with corresponding \neff\ values  are reported in Table~\ref{tab:mi_for_undersampling_regression}. For the most realistic case, with knowledge of upto 100 members ($m$) per attribute, the inference risk remains very low, with MSE values as high as $\sim5$ (\neff$<0.1)$. This risk is expected to increase with increase in membership knowledge, as in the extreme case, the adversary would have perfect knowledge of the entire training dataset. One notable exception is \boneage, where the MSE drops to $\sim0.4$ (\neff=2) for $m=500$ This is not surprising, as $m=500$ for the case of \boneage\ corresponds to $\sim15\%$ of the victim's training dataset, which is unrealistically high.

\begin{table*}[h]
\centering
\renewcommand{\arraystretch}{1.1}
\begin{tabular}{lccc}
 \toprule
{\bf Dataset/Task} & \kldabbr & MI-Based & Same Setup + \kldabbr\\
 \hline
 \censusn\ (Sex) & 56.7 (0.1) & 57.1 ($<$0.1) & 80.4 (1.8)\\
 \boneage\ (Age) & 59.1 (0.3) & 65.9 (0.3) & 50.0 ($<$0.1)\\
 \celeba\ (Male) & 53.7 ($<$0.1) & 50.0 ($<$0.1) & 64.7 (0.3)\\
 \bottomrule
\end{tabular}
\caption{Mean distinguishing accuracies (with mean \neff\ in parentheses) for the task of binary distinguishing between $\alpha_0=0.5$ and $\alpha_1$, while varying $\alpha_1$, for the standard adversary (\kldabbr), an adversary that utilizes membership inference to infer $\alpha$ for models trained with under-sampling based defense (MI-Based), and a simpler adversary that just copies the victim's re-sampling setup (Same Setup + \kldabbr).}
\label{tab:mi_for_undersampling}
\end{table*}

For the task of binary distinguishing between $\alpha_0=0.5$ and some other $\alpha_1$, it suffices to see whether the predicted ratio is sufficiently different from $0.5$. We do so by checking the predicted $\alpha$, and predict \gone($\mathcal{D}$) if it differs from $0.5$ by more than $0.03$, and \gzero($\mathcal{D}$) otherwise. As a baseline, we also consider a simpler adaptive adversary that uses the same re-sampling setup as the victim, re-sampling data for its shadow models. Such an adversary can potentially work, since the \kldtest\ compares distributions of predictions, which might be sufficiently different between re-sampled and non-sampled ($\alpha=0.5$) models. 

Mean distinguishing accuracies and corresponding \neff\ values are reported in Table~\ref{tab:mi_for_undersampling}. Cases where the adversary uses the same setup as the victim (for re-sampling) leads to significant inference leakage in most cases, with mean distinguishing accuracies as high as 80\% (\neff=1.8) for \censusn. Similarly, the MI-based distribution inference leakage is particularly high for \boneage. This is in line with previous observations with the MSE values (Table~\ref{tab:mi_for_undersampling_regression}), since the number of members corresponds to a significant portion of the victim's training data.

\end{document}